\documentclass[dvipsnames]{article}
\usepackage{colm2024_conference}

\usepackage{booktabs}
\usepackage{graphicx}
\usepackage{enumitem}
\usepackage{wrapfig}
\usepackage{algorithm}
\usepackage{algpseudocode}
\usepackage{xcolor}
\usepackage{xspace}
\usepackage{microtype}
\usepackage{amsmath}
\usepackage{colortbl}
\usepackage[utf8]{inputenc}
\definecolor{lightgray}{rgb}{0.9,0.9,0.9}
\usepackage{caption}
\usepackage{subcaption}
\usepackage{setspace}
\usepackage{url}
\usepackage{multirow}
\usepackage{colortbl}
\usepackage{tabularx}
\usepackage{blindtext}
\usepackage{pgfplots}
\pgfplotsset{compat=1.18} 
\usepackage{tikz}
\usetikzlibrary{er,positioning,bayesnet}
\usepackage{makecell}
\usepackage{tipa}
\usepackage{siunitx}
\usepackage{nicefrac}
\usepackage{tocloft}
\usepackage{listings}
\usepackage[raster,skins]{tcolorbox} %
\usepackage{xltabular}
\usepackage{adjustbox}
\usepackage{xurl}
\usepackage{rotating}
\usepackage[normalem]{ulem}
\useunder{\uline}{\ul}{}
\usepackage{amsthm}
\usepackage[utf8]{inputenc} 
\usepackage[T1]{fontenc}    
\usepackage{hyperref}               
\usepackage{amsfonts}       
\usepackage{nicefrac}       
\usepackage{microtype}      
\usepackage{subcaption} 
\usepackage{amssymb}
\usepackage{geometry}
\usepackage{babel}
\usepackage{siunitx}
\usepackage{bm}
\usepackage{bbm}
\usepackage{caption}
\usepackage{appendix}

\definecolor{purple1}{RGB}{126, 107, 196}
\definecolor{purple2}{RGB}{199, 158, 207}
\definecolor{purple3}{RGB}{214, 200, 255}
\definecolor{purple4}{RGB}{254, 240, 255}

\definecolor{darkblue}{RGB}{0, 0, 139}
\definecolor{darkgreen}{RGB}{0, 100, 0}
\definecolor{orange}{RGB}{255, 165, 0}

\newcommand{\eg}{\emph{e.g.},\xspace}
\newcommand{\ie}{\emph{i.e.},\xspace}

\newcommand\figref[1]{Fig.~\ref{#1}}

\newcommand\secref[1]{Sec.~\ref{#1}}

\newcommand\exaref[1]{Example~\ref{#1}}

\newcommand{\ours}{AgentScope\xspace}
\newcommand{\module}[1]{\textsf{#1}}

\renewcommand{\ghlink}{https://github.com/agentscope-ai/agentscope}

\newcommand*\justify{%
  \fontdimen2\font=0.4em
  \fontdimen3\font=0.2em
  \fontdimen4\font=0.1em
  \fontdimen7\font=0.1em
  \hyphenchar\font=`\-
}

\renewcommand{\texttt}[1]{%
  \begingroup
  \ttfamily
  \begingroup\lccode`~=`/\lowercase{\endgroup\def~}{/\discretionary{}{}{}}%
  \begingroup\lccode`~=`[\lowercase{\endgroup\def~}{[\discretionary{}{}{}}%
  \begingroup\lccode`~=`.\lowercase{\endgroup\def~}{.\discretionary{}{}{}}%
  \catcode`/=\active\catcode`[=\active\catcode`.=\active
  \justify\scantokens{#1\noexpand}%
  \endgroup
}

\title{AgentScope 1.0: A Developer-Centric Framework for Building Agentic Applications}

\author{
Dawei Gao,\ \   
Zitao Li,\ \   
Yuexiang Xie,\ \    
Weirui Kuang,\ \   
Liuyi Yao,\ \    
\\
Bingchen Qian,\ \    
Zhijian Ma,\ \    
Yue Cui,\ \    
Haohao Luo,\ \    
Shen Li,\ \    
Lu Yi,\ \    
Yi Yu,\ \    
\\
Shiqi He,\ \
Zhiling Luo,\ \   
Wenmeng Zhou,\ \    
Zhicheng Zhang,\ \    
Xuguang He,\ \    
\\
Ziqian Chen,\ \    
Weikai Liao,\ \    
Farruh Isakulovich Kushnazarov,\ \   
\\
Yaliang Li$^*$,\ \    
Bolin Ding$^*$,\ \    
Jingren Zhou
\\
\vspace{5mm}
\small{Alibaba Group}
}

\newcommand{\ASR}{Runtime\xspace}

\begin{document}

\maketitle

\renewcommand*{\thefootnote}{\fnsymbol{footnote}}
\footnotetext[1]{Corresponding authors, email address: \{yaliang.li, bolin.ding\}@alibaba-inc.com.} 
\renewcommand*{\thefootnote}{\arabic{footnote}}

\begin{abstract}

Driven by rapid advancements of Large Language Models (LLMs), agents are empowered to combine intrinsic knowledge with dynamic tool use, greatly enhancing their capacity to address real-world tasks. 
In line with such an evolution, \ours introduces major improvements in a new version (1.0), towards comprehensively supporting flexible and efficient tool-based agent-environment interactions for building agentic applications. 

Specifically, we abstract foundational components essential for agentic applications and provide unified interfaces and extensible modules, enabling developers to easily leverage the latest progress, such as new models and MCPs. 
Furthermore, we ground agent behaviors in the ReAct paradigm and offer advanced agent-level infrastructure based on a systematic asynchronous design, which enriches both human-agent and agent-agent interaction patterns while improving execution efficiency. Building on this foundation, we integrate several built-in agents tailored to specific practical scenarios.
\ours also includes robust engineering support for developer-friendly experiences. We provide a scalable evaluation module with a visual studio interface, making the development of long-trajectory agentic applications more manageable and easier to trace. 
In addition, \ours offers a runtime sandbox to ensure safe agent execution and facilitates rapid deployment in production environments. With these enhancements, \ours provides a practical foundation for building scalable, adaptive, and effective agentic applications.

\end{abstract}

\section{Introduction}
\label{sec:introduction}

The rapid advancement of Large Language Models (LLMs)~\citep{gpt4, claude, llama4,yang2025qwen3,team2025kimi} has led to remarkable progress in artificial intelligence. 
A key feature of modern LLMs is their ability to call and interact with external tools~\citep{gpt4, hurst2024gpt, claude, llama4, yang2025qwen3, team2025kimi}, greatly enhancing their functional scope. 
This tool-calling capability allows LLMs to automatically process external databases, execute computational tasks, and interact with different APIs, thereby extending their utility beyond intrinsic reasoning and language processing. 

Such advancements have laid a robust foundation for developing powerful LLM-based agent applications that can effectively interface with the world through a variety of tools, to perform diverse and complex tasks with increased autonomy and precision~\citep{qin2024toolllm, toollearning, zhang2024large, cui25enhancing, yuan2024easytool}.
By interacting with the environment, LLM-based agents have demonstrated immense potential in a wide range of applications~\citep{metagpt, simulation, langchain}, proving increasingly capable of solving complex real-world problems while supporting flexible interactions with both users and environments.

Following this trend, the focus of LLM-based agent frameworks has shifted from relying solely on intrinsic reasoning to empowering agents to perceive and interact with environments via an array of tools. 
Consequently, building flexible and efficient agent frameworks that support tool-based perception and interaction has emerged as a promising direction in both academic research and industrial practice~\citep{autogen, agno2024, langchain}.

\begin{figure}
    \centering
    \includegraphics[width=0.8\linewidth]{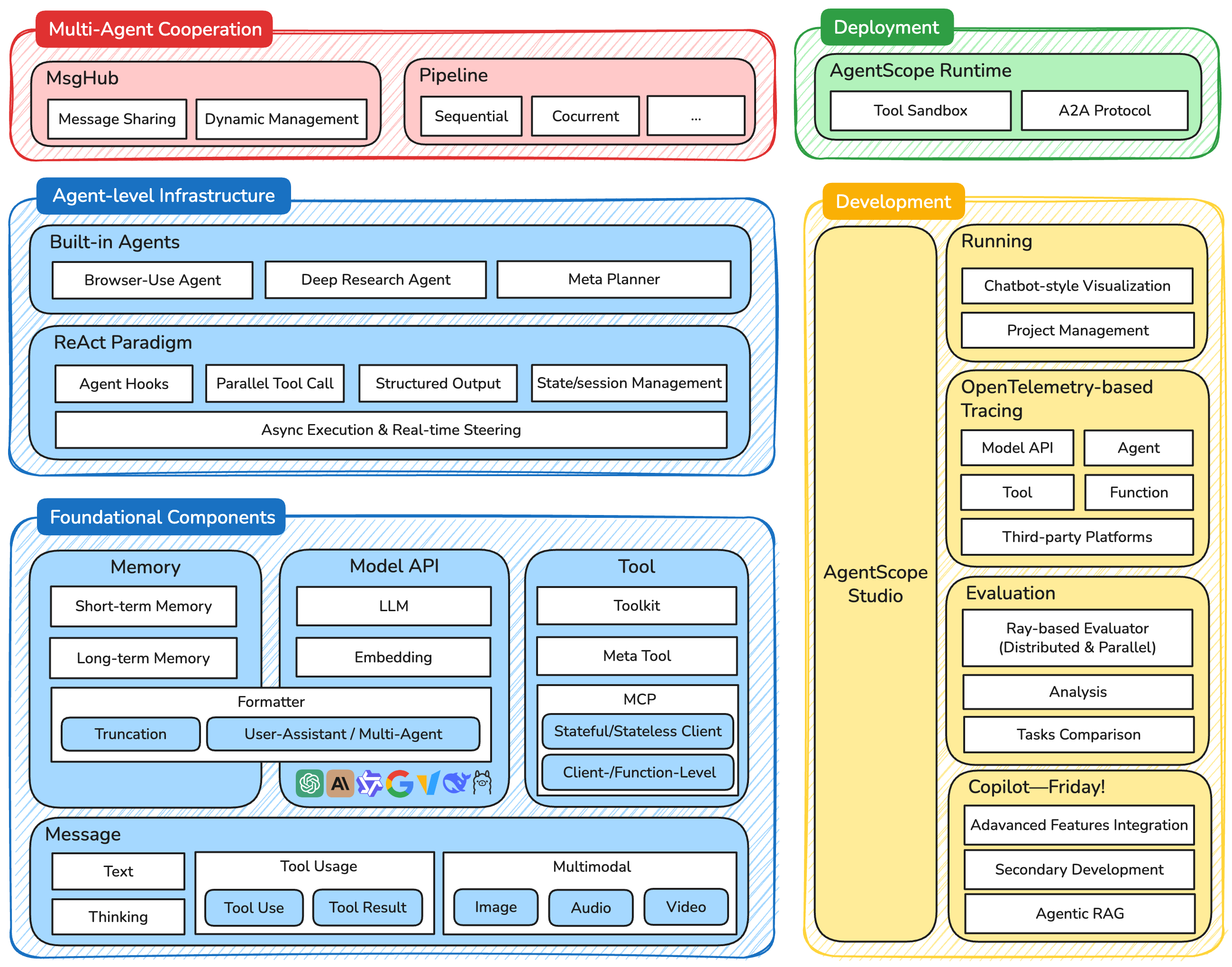}
    \caption{The overview of \ours framework.}
    \label{fig:framework}
\end{figure}

Motivated by these insights and evolving demands, we introduce a new version of \ours with a novel architecture grounded in the ReAct~\citep{react} paradigm. 
This paradigm combines explicit reasoning with actions, enabling agents to analyze tasks, call tools, observe execution results, and iteratively refine their steps in a closed loop. The overall architecture of \ours is illustrated in \figref{fig:framework}. To maximize flexibility and usability, \ours incorporates several design choices that allow developers to assemble, adapt, and extend agentic applications for real-world settings.

(a) \textbf{Foundational Components}. At the core of \ours is a set of foundational components that make building agentic applications both straightforward and flexible. We abstract the required components into four modules: \module{message}, \module{model}, \module{memory}, and \module{tool}. Our design emphasizes strong modular decoupling, broad compatibility across different application types, and extensibility for downstream customization. 
For example, multimodal information can be uniformly formatted as messages for transmission among agents, and diverse Model Context Protocols (MCPs)~\citep{MCP} can be registered as tools to enrich how agents interact with environments.
These foundational components can be composed flexibly to serve a broad set of practical applications.

(b) \textbf{Agent-level Infrastructure}. \ours adopts the ReAct paradigm as the primary and recommended agent architecture, as it provides a simple yet effective paradigm for agent-environment interaction. 
Building upon this, \ours natively supports parallel tool calls, asynchronous executions, and real-time steering, delivering industrial-grade performance and efficiency for running agentic applications.
Additionally, \ours integrates several built-in agents, including a browser-use agent, a deep research agent, and a meta-planner agent. 
These agents are built on the basic ReAct agent and equipped with task-specific tools, hook functions, and prompts to address representative and well-studied scenarios~\citep{deepresearchagent, browseragent}. 
Developers can use these agents out of the box or treat them as starting points for further customization.

(c) \textbf{Developer-friendly Experiences}.
To provide developer-friendly experiences throughout all stages of development and deployment, \ours integrates a comprehensive suite of toolkits designed to streamline the entire workflow. The evaluation module offers a unified interface for assessing agent performance and includes two specialized evaluators, enabling users to flexibly balance debugging convenience with computational efficiency. Besides, Studio, a graphical interface for process monitoring and result tracing, supports multi-granularity and multi-dimensional analysis of running trajectories and evaluation results.
A runtime sandbox allows developers to easily configure and launch agent execution and deployment environments according to specific tool requirements.
These toolkits ensure that \ours delivers a smooth, efficient, and developer-friendly experience.

\noindent\textbf{Roadmap.}
In this manuscript, the details of the foundational agentic components will be introduced in the following \secref{sec:foundational_components}, including message, model interface, memory and tool for agents. 
The built-in ReAct-based agent-level functionalities will be elaborated in \secref{sec:agent_infra}.
The engineering support modules, including the evaluation module, studio, and runtime sandbox, are illustrated in \secref{sec:developer}.
Last but not least, we present some examples and applications in \secref{sec:application} to demonstrate the potential of \ours.
\section{Foundational Components}
\label{sec:foundational_components}

In this section, we introduce the foundational components in \ours, including \module{message}, \module{model}, \module{memory}, and \module{tool} modules. 
For each component, we present its design goals and principles, implementation details, and illustrative examples for a better understanding.
\subsection{Message}
\label{subsec:message}

The \module{message} module is the basic data unit in \ours, which enables information exchange among agents, presentation in the user interface, and storage in memory.
Meanwhile, it serves as the unified information abstraction and medium between \ours and different LLM APIs.

A message object (\ie \texttt{Msg}) comprises the following key fields:

\begin{itemize}
    \item {\it Name}: Records the name of the sender that produced the message, distinguish agents in multi-agent applications.
    \item {\it Role}: Indicates the role of sender, which can be one of "user", "assistant", or "system".
    \item {\it Content}: Contains the main payload of the message. 
    It can be a simple text string or a sequence of structured \texttt{ContentBlock} objects, such as text blocks, image blocks, audio blocks, video blocks, tool usage blocks, tool results blocks, and thinking blocks. The design of \texttt{ContentBlock} enables agents to exchange multimodal content, tool-usage details, and reasoning information, thereby natively supporting a range of practical agentic applications.
    \item {\it Metadata}: Provides an option to attach additional meta information to the message, such as structured outputs. 
\end{itemize}

In addition, each message is automatically assigned a {\it timestamp} and a unique {\it id} upon instantiation to ensure traceability.
Example~\ref{example:msg} shows how to create messages in AgentScope.

\begin{lstlisting}[language=Python, caption=Example of message creation in AgentScope, label=example:msg, float=t]
from agentscope.message import Msg, ToolUseBlock

# Example 1: Create a Textual Message
textual_msg = Msg(
    name="Jarvis",
    role="assistant",
    content="Hello! How can I help you?",
)

# Example 2: Create a Tool Use Message 
msg_tool_call = Msg(
    name="Jarvis",
    role="assistant",
    content=[
        ToolUseBlock(
            type="tool_use",
            id="xxx",
            name="get_weather",
            input={"location": "Beijing"}
        )
    ]
)
\end{lstlisting}

\subsection{Model}
The \module{model} module provides a unified abstraction for integrating diverse LLM APIs, enabling seamless interoperability across model providers while delivering a consistent interface and functionality. 
Such an abstraction and design philosophy address the inherent heterogeneity among different model providers, who might use different API specifications, parameter formats, and response structures. 
AgentScope integrates a wide range of LLM providers with full feature compatibility, as summarized in Table~\ref{tb:models}.

\begin{table}[t]
    \centering
    \caption{The integrated LLM providers and their features in \ours.}
    \begin{tabular}{llccccc}
        \toprule
        \textbf{Provider} & \textbf{Class} & \textbf{Streaming} & \textbf{Tools} & \textbf{Vision} & \textbf{Reasoning}\\
        \midrule
        OpenAI, DeepSeek, vLLM & \texttt{OpenAIChatModel} & \checkmark & \checkmark & \checkmark & \checkmark  \\
        DashScope & \texttt{DashScopeChatModel} & \checkmark & \checkmark & \checkmark & \checkmark  \\
        Anthropic & \texttt{AnthropicChatModel} & \checkmark & \checkmark & \checkmark & \checkmark  \\
        Gemini & \texttt{GeminiChatModel} & \checkmark & \checkmark & \checkmark & \checkmark  \\
        Ollama & \texttt{OllamaChatModel} & \checkmark & \checkmark & \checkmark & \checkmark  \\
        \bottomrule
    \end{tabular}
    \label{tb:models}
\end{table}

Built on the \texttt{ChatModelBase} abstract class, different model implementations share a unified and standardized interface that includes (a) model‑specific formatters, (b) asynchronous model calls, (c) a unified response schema, and (d) usage tracking and hook functions.
More details are provided in the rest of this subsection.

\paragraph{Model-specific Formatters.}
Different model APIs set their own requirements for the inputs to LLMs, often differing subtly in the input formats, role specifications, and content structures.
To bridge the gap between the message in \ours and the heterogeneous input format of different LLM APIs, we develop an abstract \texttt{format} method in the \texttt{FormatterBase} class to transform \texttt{Message} objects into provider-specific data structures.
We provide two specialized formatters for each model provider, including a \texttt{ChatFormatter} for supporting single-agent interactions, and a \texttt{MultiAgentFormatter} for handling multi-participant conversations where speaker identification and role management are crucial. 
Considering that not all model providers support multi-agent messages natively, \texttt{MultiAgentFormatter} utilizes conversation history prompts and structured content to ensure compatibility with standard chat completion endpoints.

This module also unifies the processing of multimodal content, which automatically converts local media (\eg images and audio) to base64 format when required, and preserves URL references according to the provider's specific requirements.
As a result, developers can handle multimodal inputs seamlessly across different model providers without additional application-level format management.

\paragraph{Asynchronous Model Calls.} 
The input of a \texttt{model} object includes messages, tools, and other parameters supported by the LLM APIs. The messages parameter carries the conversation history as a list of dictionaries produced by the corresponding formatter, while the tools parameter is a set of JSON schemas describing available tool functions. Besides, for some LLMs, the optional \texttt{tool\_choice} parameter controls the tool selection strategy. 

AgentScope natively supports asynchronous model calls, providing a non-blocking design and efficient streaming response via Python’s asynchronous generators. With streaming disabled, the calling method returns a single \texttt{ChatResponse} object containing the complete model output. With streaming enabled, it returns an asynchronous generator that yields \texttt{ChatResponse} updates in real time as the model produces content, following a cumulative scheme that each chunk includes all content generated so far.

\paragraph{A Unified Response Schema.}
In \ours, model responses are encapsulated in the \texttt{ChatResponse} dataclass, which abstracts provider-specific output formats into a unified schema. 
Specifically, a model response exposes a \texttt{content} field that supports heterogeneous content types, including \texttt{TextBlock} for textual responses, \texttt{ToolUseBlock} for function calls, and \texttt{ThinkingBlock} for reasoning traces. 
Additional metadata includes unique identifiers, creation timestamps, and usage statistics of input tokens, output tokens, and processing time for monitoring and analysis.

The unified response schema enables sophisticated reasoning outputs across multiple providers. We use the \texttt{ThinkingBlock} objects to expose internal reasoning traces, with support for models from OpenAI, Anthropic, Gemini, and Ollama that offer explicit reasoning capabilities. 
\ours also provides fine-grained control over reasoning output via provider-specific mechanisms. For example, OpenAI’s o-series models support reasoning effort levels (\texttt{"low"}, \texttt{"medium"}, and \texttt{"high"}), while Anthropic and Gemini expose configurable token budgets for reasoning processes. This abstraction allows developers to leverage advanced reasoning across providers while consuming a consistent response schema, regardless of the specific implementations.

\paragraph{Usage Tracking and Hook Functions.}
The \texttt{ChatUsage} object provides fine-grained monitoring of model consumption through comprehensive metrics covering input tokens, output tokens, and processing latency.
This unified tracking module enables per-invocation resource accounting across providers, supporting detailed cost analysis, comparative efficiency studies, and the implementation of usage-based billing and rate-limiting mechanisms in production. 
Its standardized format allows developers to build provider-agnostic cost dashboards and automated budget controls without vendor-specific integrations.

\ours offers comprehensive extensibility via a multi-layer hook system for deep integration with enterprise monitoring and observability stacks. It includes built-in distributed tracing through a designed \texttt{@trace\_llm} decorator, which automatically instruments model calls with OpenTelemetry-compatible~\citep{OpenTelemetry} spans that capture request parameters, response metadata, token-usage statistics, and error conditions. These traces integrate seamlessly with systems such as Arize-Phoenix~\citep{Arize-Phoenix} and Langfuse~\citep{Langfuse}.

\subsection{Memory}
\label{sec:memory}

The \module{memory} module is designed to provide contextual information for subsequent reasoning and action steps, including conversation history, execution trajectories, and cross-conversation data such as user preferences. 
In \ours, the \module{memory} module consists of both short-term and long-term memory components.

\subsubsection{Short-term Memory}
Short-term memory is essential for agents to keep track of recent communications and execution trajectories. 
In \ours, \texttt{InMemoryMemory} serves as the default buffer for storing this information. The implementation maintains an in-memory list of \texttt{Msg} objects, capturing the complete communication context between agents and users, as well as tool execution trajectories.

The \texttt{InMemoryMemory} class provides basic operations for memory management, including adding new messages to the dialogue history, retrieving a range of memory content, deleting specific messages by index, and clearing the entire memory buffer. During agent execution, particularly within reasoning-acting loops, the memory is automatically updated. The incoming messages are promptly added before processing, while the responses and tool usage records are stored to preserve the full interaction trajectory.

The design of short-term memory in \ours ensures agents maintain contextual awareness throughout multi-step executions and multi-turn conversations, while offering the flexibility to manage memory size and content according to different application needs.

\subsubsection{Long-term Memory}
Long-term memory provides a structured mechanism for persistent context management, enabling agents to retain and leverage information across conversations, such as user preferences, task history, and interaction patterns. 

\paragraph{Design and Abstraction.}
The abstract class \texttt{LongTermMemoryBase} serves as the core abstraction for all long-term memory implementations within \ours, which defines a standardized protocol for memory operations for ensuring consistency across different backends and use cases.

The abstract class specifies four key methods, organized into two distinct operational paradigms:

\begin{itemize}
    \item \textit{Developer-Controlled Methods}:
\begin{itemize}
\item \texttt{record}: Records structured information from message sequences, typically invoked at predefined stages in the agent workflow (\eg session start or end).
\item \texttt{retrieve}: Retrieves relevant memory entries based on the content of input messages, enabling context-aware responses.
\end{itemize}

\item \textit{Agent-Controlled Methods}:
\begin{itemize}
\item \texttt{record\_to\_memory}: Allows the agent to autonomously store information it deems important during reasoning.
\item \texttt{retrieve\_from\_memory}: Enables the agent to perform keyword-driven queries to retrieve specific knowledge when needed.
\end{itemize}
\end{itemize}

This dual-paradigm design supports flexible memory management strategies. Developer-controlled methods ensure reliable and systematic memory operations at critical points in the agent lifecycle, while agent-controlled methods are automatically registered in the agent’s toolkit, empowering the agent to make context-sensitive decisions about memory usage during execution.

\textbf{A Specific Implementation.}
The \texttt{Mem0LongTermMemory} class provides a specific implementation of long-term memory based on the mem0 library~\citep{mem0}, demonstrating how external memory systems can be integrated into \ours while maintaining the framework’s interface and control mechanisms.

By inheriting from \texttt{LongTermMemoryBase}, \texttt{Mem0LongTermMemory} implements all memory methods and leverages advanced capabilities in Mem0, such as semantic indexing, retrieval, and memory evolution. 
To support diverse deployment scenarios, two configuration strategies are provided:
\begin{itemize}
\item \textit{Individual Parameter Configuration}: If no \texttt{mem0\_config} is supplied, the class constructs its configuration from individual parameters. Explicit specification of \texttt{model} and \texttt{embedding\_model} is required for proper initialization.
\item \textit{Pre-configured Mem0 Configuration}: Developers familiar with Mem0 can pass a pre-defined \texttt{mem0\_config} object. Individual parameters may be used to override specific settings, enabling fine-grained customization.
\end{itemize}
These strategies ensure accessibility for new users and flexibility for advanced users, promoting seamless integration across a wide range of applications.

In this way, the long-term memory in \ours provides a comprehensive and extensible solution for persistent knowledge management. Through its abstract base class design, support for multiple control modes, and pluggable backend implementations, this module accommodates both systematic and opportunistic memory usage patterns.

\subsection{Tool}
\label{sec:tool}

\begin{table}[t]
    \centering
    \footnotesize
    \caption{The provided interfaces in the \texttt{Toolkit} module.}
    \begin{tabular}{lll}
        \toprule
        \textbf{Type} & \textbf{Interfaces} & \textbf{Descriptions}   \\
        \midrule
        \multirow{4}{*}{Basic usage} & register\_tool\_function & Register a target function \\
                                     & execute\_tool\_function & Execute the function call \\
                                     & remove\_tool\_function & Remove tool function from the toolkit by its name \\                                    
                                     & get\_json\_schemas & Get the JSON schema of tools\\
        \midrule
        \multirow{2}{*}{MCP-related} & register\_mcp\_client & Register tool functions from an MCP client \\
                                     & remove\_mcp\_clients & Remove tool functions from the specific MCP clients \\
        \midrule
        \multirow{3}{*}{Group-wise Management} & create\_tool\_group & Create a tool group within the toolkit \\
        & update\_tool\_groups & Update the activation status of the given tool groups \\
        & remove\_tool\_groups & Remove tool functions from the toolkit by their group names \\
        \bottomrule
    \end{tabular}
    \label{tab:toolkit}
\end{table}

\begin{figure}[t]
    \centering
    \includegraphics[width=\linewidth]{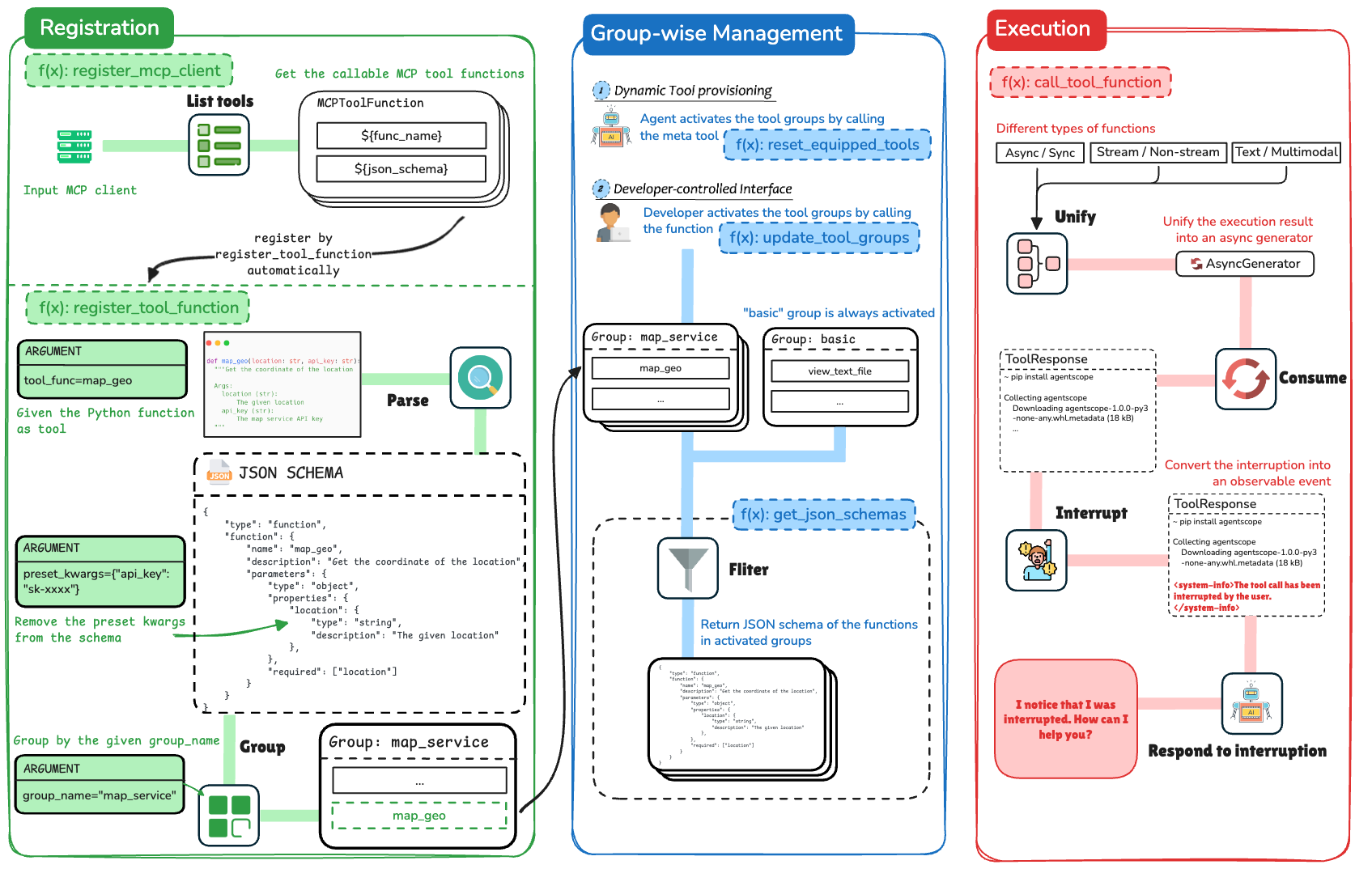}
    \caption{The usage of the {\tt Toolkit} module in \ours, including tools registration (green), group-wise management (blue), and execution (red).}
    \label{fig:toolkit}
\end{figure}

\ours accommodates a wide range of callable objects as tools, including various functions and MCPs. We define a \texttt{Toolkit}, as the core of the \module{tool} module, to achieve flexible tool management by standardizing tool definitions into JSON schema and providing unified interfaces for their registration and execution.
The interfaces of \texttt{Toolkit} are summarized in Table~\ref{tab:toolkit}, and the usage of \texttt{Toolkit} is illustrated in \figref{fig:toolkit}.

\subsubsection{Tool Registration and Execution}  
\label{subsec:tool_register_execution}

Tool registration in \texttt{Toolkit} is centered on the \texttt{register\_tool\_function} interface. 
In addition to adding and maintaining the tool function for future usage, this interface is primarily responsible for preparing JSON schema for the tool function, which is essential for LLMs to accurately interpret tool functions and invoke them at appropriate times. 
When JSON schemas are not explicitly provided, \texttt{Toolkit} automatically constructs one with information from the function docstring, allowing developers to register tool functions with minimal effort.

Besides, the \texttt{register\_tool\_function} interface is highly extensible. Developers can attach preset arguments (\eg API keys and credentials), define post-processing logic to refine raw outputs, and dynamically extend the tool schema using a \texttt{BaseModel} in Pydantic~\citep{Pydantic}. Such extensibility is particularly useful for implementing complex interaction patterns, \eg developers can programmatically add a "thinking" parameter to all tools to enable Chain-of-Thought (CoT) reasoning~\citep{cot}.

As for tool execution, the \texttt{call\_tool\_function} interface abstracts away the inherent complexity of handling various tool outputs by unifying the outputs of all registered functions, whether synchronous or asynchronous, streaming or non-streaming, into a consistent asynchronous generator. This design allows developers to invoke different tools through a unified interface, simplifying the efforts required to process diverse outputs.

It is worth noting that we enhance the robustness of the provided asynchronous generator, especially for running interactive and long-running tasks. If the execution of a streaming tool is interrupted (\eg by an \texttt{asyncio} cancellation event), the toolkit gracefully preserves all results yielded up to that point and appends a clear system notification, such as \textit{``tool execution was interrupted''}, to the output stream. This mechanism ensures that partial progress is retained and provides explicit context regarding interruptions, which is crucial for building resilient and user-responsive agents.

\subsubsection{Fine-grained MCP Management}

Integrating remote services via MCP is a common demand in agentic applications~\citep{hou2025model,deepresearchagent,toollearning,wang2024survey,luo2025large}. 
However, raw remote functions often necessitate client-side adaptations, such as result post-processing, parameter filtering, or composition into more complex workflows. 
To tackle this, \ours provides an advanced MCP client architecture that enables fine-grained management of remote tools at both the client and function levels.

\begin{figure}[th]
    \centering
        \begin{subfigure}{0.49\linewidth}
        \centering
        \includegraphics[width=\linewidth]{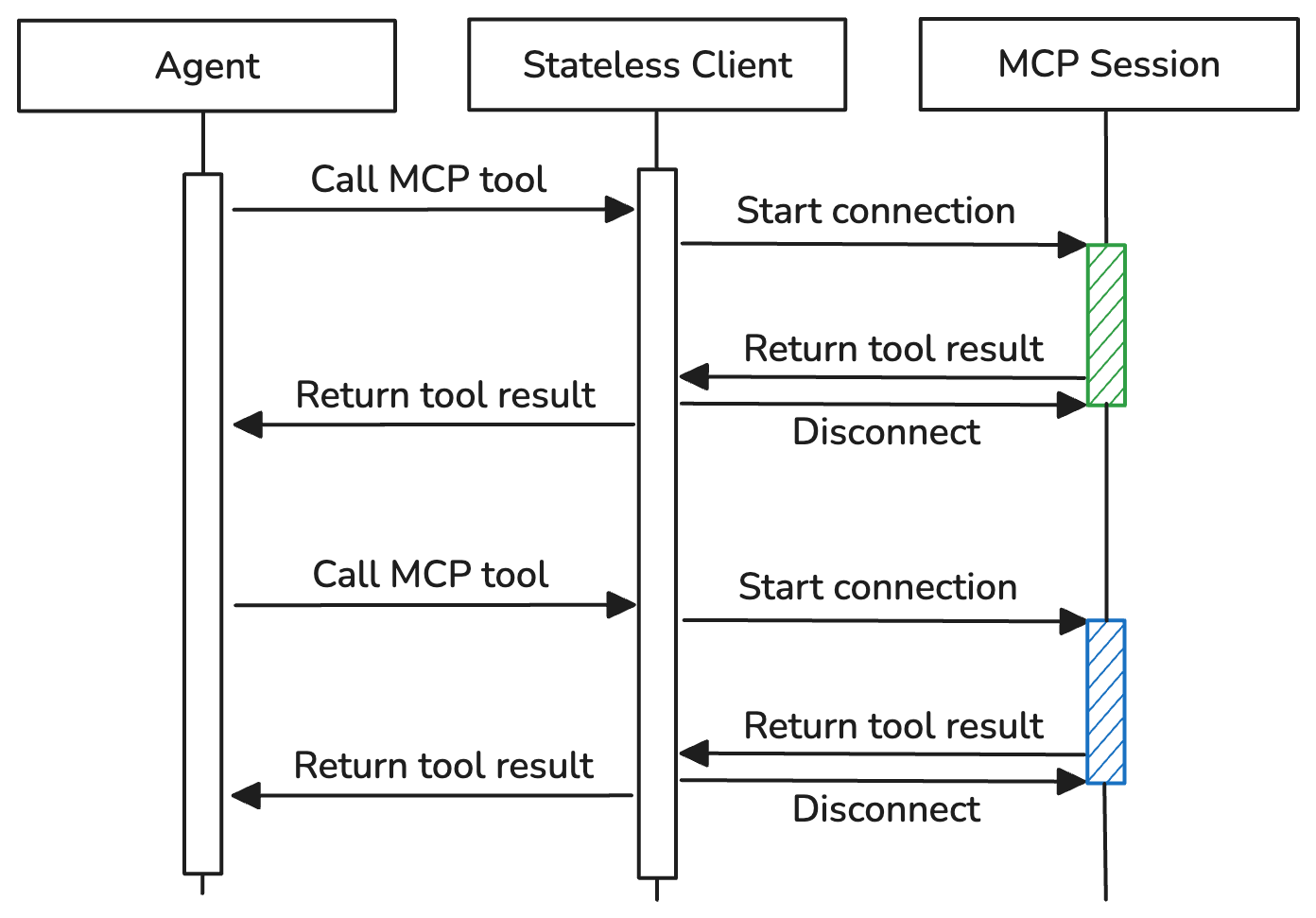}
        \caption{Stateless MCP client}
        \label{fig:stateless_client}
    \end{subfigure}
    \centering
        \begin{subfigure}{0.49\linewidth}
        \centering
        \includegraphics[width=\linewidth]{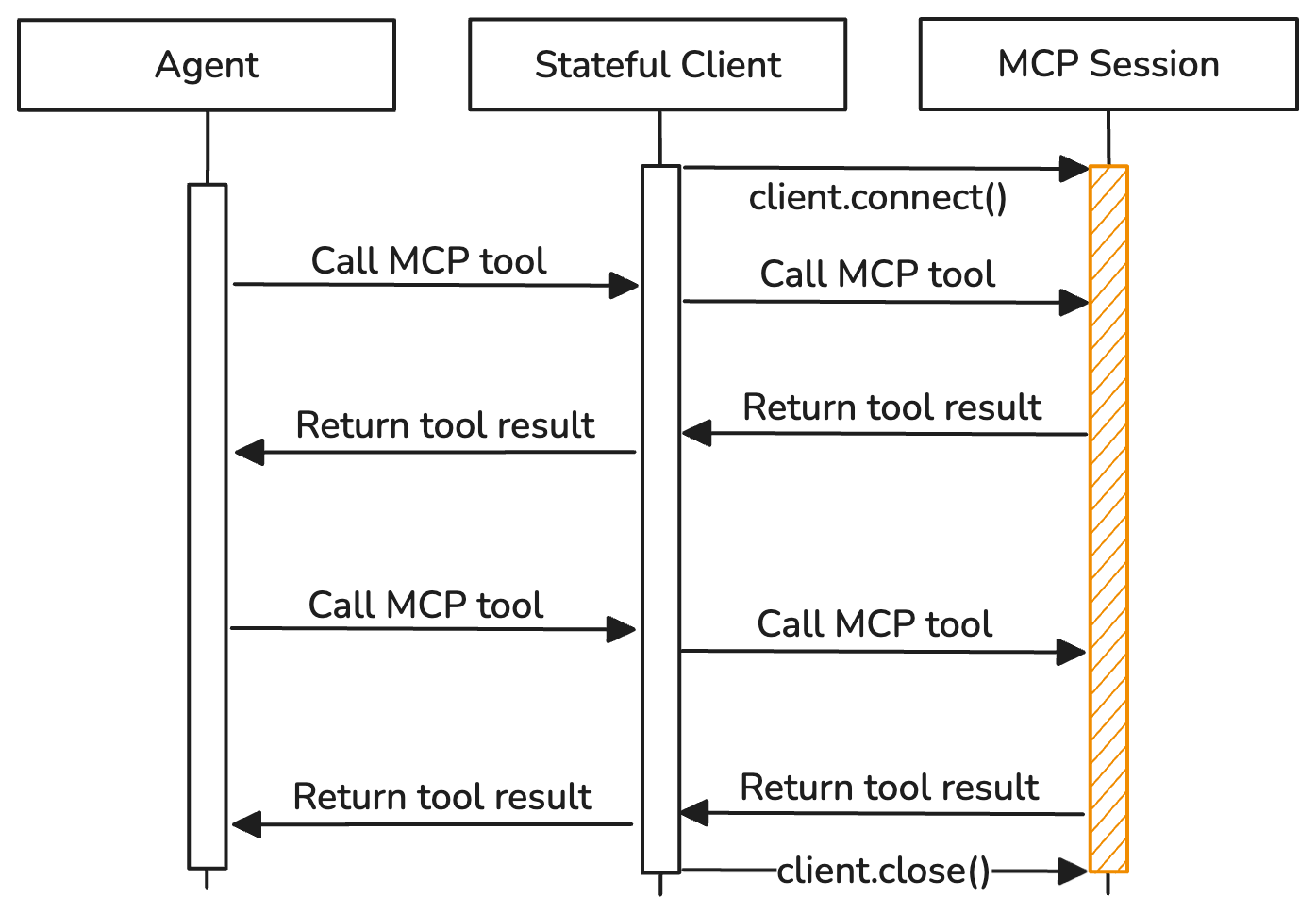}
        \caption{Stateful MCP client}
        \label{fig:stateful_client}
    \end{subfigure}
    \caption{The sequence diagram of the stateless and stateful MCP clients. The stateless client (Left) establishes a new session per tool call, while the stateful client (Right) maintains a persistent connection.}
    \label{fig:mcp_clients}
\end{figure}

\paragraph{Stateful and Stateless Clients.}
Central to our MCP client architecture is a dual-client design, providing both \textit{stateful} and \textit{stateless} clients to accommodate different interaction patterns. 
The choice between them depends on session management requirements. Specifically, a stateful client establishes a persistent connection to an MCP server via explicit \texttt{connect} and \texttt{close} interfaces. This design ensures that all subsequent tool calls occur within the same session, making it feasible for services where state continuity is essential, such as a remote browser session that must maintain cookies and context across multiple actions. 
In contrast, a stateless client follows an ephemeral connection model. It automatically establishes a connection immediately before a tool call and terminates it right after, thereby minimizing resource overhead. This approach is well-suited for lightweight and transactional services that do not depend on session state. The distinct lifecycle management of these two clients is illustrated in \figref{fig:mcp_clients}.

\paragraph{Client-Side Tool Abstraction.}
Our MCP clients provide a powerful abstraction that transforms remote endpoints into native and first-class tools. Instead of simply listing function names, the client generates local callable objects that serve as proxies for their remote counterparts. These proxy objects can be directly registered with \texttt{register\_tool\_function}, rendering remote services indistinguishable from local ones from the agent's perspective.

This seamless integration is crucial for enabling advanced customization. Since these proxies behave as standard Python objects, developers can easily wrap them in new functions to implement bespoke logic. For example, developers can construct a composite function that first invokes a remote search tool, then uses a local regular expression to filter the results before passing them to another remote summarization tool. Such composability allows developers to adapt and combine raw remote services into high-level and task-specific tools without requiring server-side modifications, greatly enhancing the agent's flexibility and capability.

\subsubsection{Group-Wise Tool Management}
\label{subsec:group_wise_tool}
As the number of integrated tools increases, agents encounter a "paradox of choice". 
Recent studies have shown that an overabundance of tools can actually degrade performance, leading to failures in selecting the appropriate tool or configuring its parameters correctly~\citep{paramanayakam2025less,liu2024toolace}. This challenge not only increases the cognitive load on the agent but also consumes valuable context length with redundant tool descriptions.

To tackle this, \ours introduces a group-wise tool management strategy. This design is motivated by the observation that many tools are naturally utilized within task-oriented workflows. For example, a web automation task typically involves a sequence of related actions such as navigating to a URL, clicking web elements, and entering text. Rather than presenting these tools as isolated options, grouping them provides a more structured and efficient approach.

For implementation, we provide several interfaces in \texttt{Toolkit}. Developers can use \texttt{create\_tool\_group} to logically bundle related tools, such as creating a "browser tools" group for all web-related functions. Subsequently, the \texttt{update\_tool\_groups} interface allows for dynamic activation or deactivation of the entire tool set. This mechanism enables an agent to operate with a streamlined and context-aware subset of its full capabilities at any given moment. For example, when the agent needs to perform web browsing, it can activate the "browser tools" group, making only the relevant tools available.

Such a lightweight and flexible strategy significantly reduces the search space for tool selection, thereby improving the agent's efficiency and reliability.

\section{Agent-level Infrastructure}
In this section, we provide details on the agent-level infrastructure of \ours. We discuss our design principles and highlight the features that make this framework effective for real-world agentic applications. We adopt the ReAct~\citep{react} paradigm as the primary and recommended agent architecture, enabling flexible agent-environment interactions. Building on this, we integrate several built-in agents tailored for specific practical scenarios. Besides, we also introduce how to construct multi-agent applications in \ours.

\label{sec:agent_infra}
\subsection{Architecture Based on the ReAct Paradigm}
\label{sec:agent}

\subsubsection{Overview}

\begin{figure}
    \centering
    \includegraphics[width=0.8\linewidth]{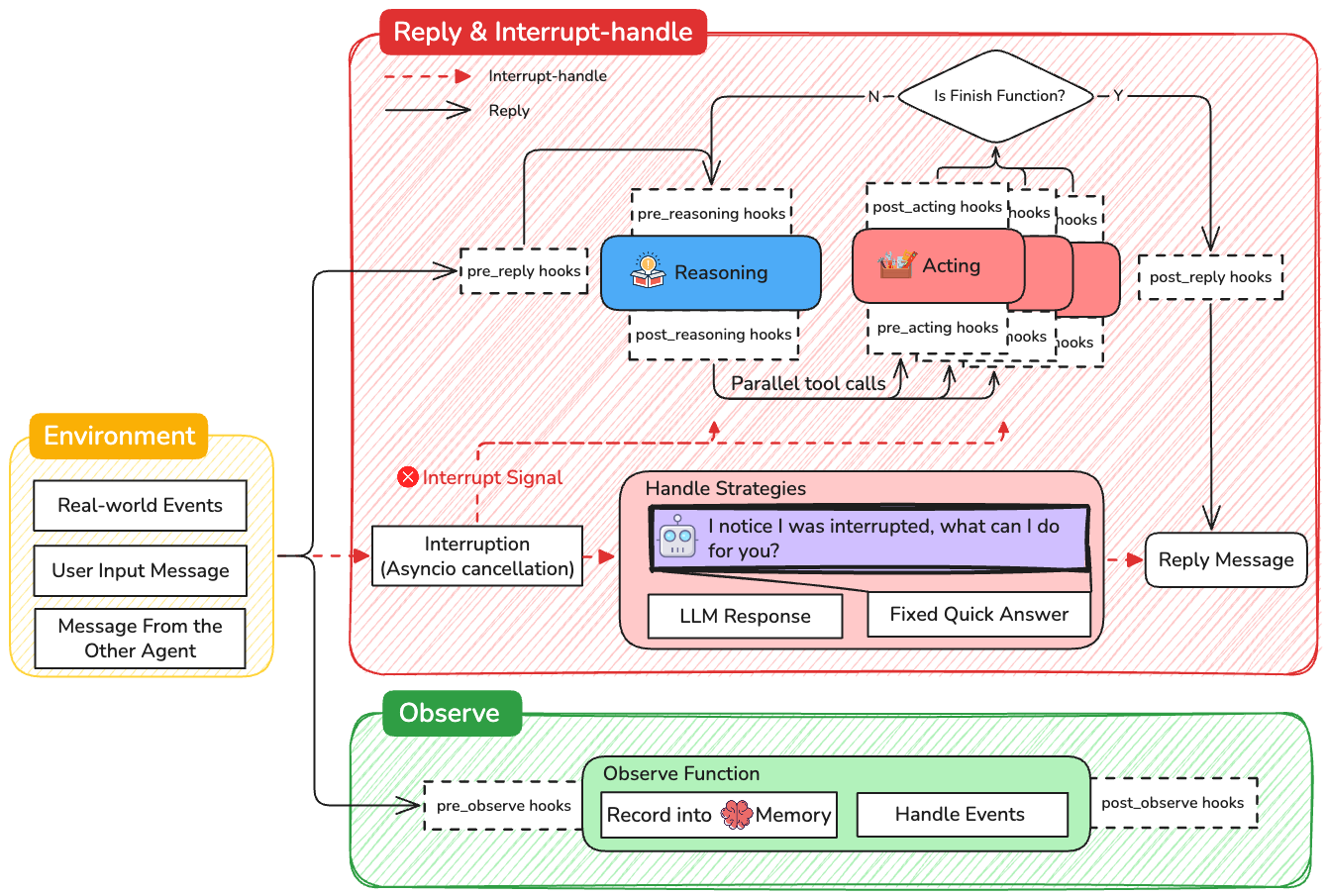}
    \caption{The workflow of the ReAct agent in \ours.}
    \label{fig:react_agent}
\end{figure}

The ReAct paradigm~\citep{react} combines reasoning with actions, providing a simple yet effective paradigm for agent-environment interaction. 
In \ours, we adopt the ReAct paradigm as the primary and recommended agent architecture, steering towards a truly application-oriented framework. 

In \ours, an agent is designed as an entity that interacts with its environment through well-defined interfaces, encompassing three core functionalities as shown in \figref{fig:react_agent}:

\begin{itemize}
    \item \textit{Reply}: This function serves as the agent’s primary active response mechanism. When receiving a user query, the agent employs this function to perform reasoning, take actions, and generate conclusive responses.
    \item \textit{Observe}: This function enables the agent to process external information, such as environmental changes or broadcast messages, and update its internal state or memory. Applying this function would not produce a response to users.
    \item \textit{Handle Interrupt}: To support seamless human-agent collaboration, this function provides a means of handling interruptions. Triggered by external signals, it allows the agent to pause ongoing operations and react promptly to interruptions (\eg urgent requests from the user).
\end{itemize}

The intelligence driving the reply function is powered by the ReAct paradigm. Specifically, the agent initiates an iterative loop of reasoning and acting once receiving a user query. This loop continues until the agent reaches a conclusion and generates a response. In each reasoning-acting cycle, the agent first produces a thought to plan its next step, and then performs an action (\eg calling a tool) to interact with the environment and gather action results.

While this loop forms the cognitive core of the agent, building an effective framework for real-world applications requires significant effort.
Rather than limiting the framework to a minimal implementation, \ours is equipped with a comprehensive suite of features designed to deliver the following key advancements:

\begin{itemize}
\item \textit{Advanced Interactivity}: We enable fluid, real-time collaboration by allowing users to interrupt and steer the agent's reasoning process.
\item \textit{Operational Flexibility and Efficiency}: We extend the agent's tool-using capabilities beyond sequential actions, supporting dynamic, task-aware tool selection and parallel execution.
\item \textit{Engineering Robustness and Extensibility}: We provide foundational mechanisms for automated state persistence and non-invasive customization, ensuring the framework is deployable, adaptable, and easy to debug.
\end{itemize}

In the rest of this subsection, we provide details of the specific features that enable these advancements.

\subsubsection{Real-time Steering}
\label{sec:realtime_steering}

Real-time steering empowers users to guide, correct, or redirect the agent during task execution, transforming interactions from rigid and monolithic processes into flexible and collaborative experiences. \ours achieves real-time steering by gracefully pausing the ongoing ReAct loop upon receiving an external interruption signal, utilizing \texttt{asyncio} cancellation as the underlying mechanism.

Developers can implement various handling strategies in the \texttt{handle\_interrupt} method to define how the agent reacts to interruptions. For example, the agent can return a quick response or invoke the LLMs to generate a context-aware reaction.

A key innovation in our design is treating interruptions not merely as control signals, but as observable events. Therefore, the agent can capture the context of each interruption and integrate it into its state. For example, a partial LLM response or a preempted tool output can be preserved in the agent’s memory, with an annotation to indicate the user interruption. This design enables the agent to maintain contextual awareness of interruptions, informing its subsequent reasoning and decisions about whether to resume, revise, or alter its course of action.

\subsubsection{Parallel Tool Calling and Dynamic Tool Provisioning}

To achieve operational flexibility and efficiency, we move beyond the standard sequential tool-use paradigm by enhancing agents with parallel tool calling and dynamic tool provisioning capabilities.

\paragraph{Parallel Tool Calling.}  
To improve efficiency, agents are allowed to generate multiple tool calls within a single reasoning step, and these tool calls can be executed in parallel, as introduced in \secref{subsec:tool_register_execution}. This parallel approach reduces task latency compared to a sequential execution, and is particularly effective for I/O-bound tasks. The process involves two steps: (a) The LLM is prompted to generate several concurrent tool calls; (b) These calls are dispatched for parallel execution using \texttt{asyncio.gather}. Then these action results are aggregated as observations for the agent's next reasoning step.

\paragraph{Dynamic Tool Provisioning.}  
To provide functional adaptability, we introduce a mechanism for \textit{dynamic tool provisioning} in \ours, centered around the \texttt{reset\_equipped\_tools} function. This function serves as a callable tool for agents, enabling them to autonomously modify their available tool set during task execution, drawing on the group-wise tool management framework introduced in \secref{subsec:group_wise_tool}.

Specifically, whenever deemed necessary, the agent can use \texttt{reset\_equipped\_tools} to activate or deactivate certain groups of tools by specifying the group name. This empowers a single agent to seamlessly handle complex and multi-stage workflows, \eg starting with a "web-browsing" tool set for research, and later switching to a "programming" tool set for implementation.

In this way, the agent can tailor its capabilities to the specific stage of the task, rather than being constrained by a predefined, one-size-fits-all tool set. Meanwhile, by limiting the available tools to those relevant for the current phase, the approach reduces the complexity of agent action selection and conserves valuable context window space.

\subsubsection{State Persistence and Non-Invasive Customization}

To enhance robustness and extensibility, \ours incorporates two novel mechanisms: an automated system for state persistence and a flexible interface for non-invasive customization.

\paragraph{State Persistence.}
We implement an automated and compositional state management system through the \texttt{StateModule} base class, which supports dual-mode registration. Firstly, attributes of any \texttt{StateModule} instance that are themselves \texttt{StateModule} objects are automatically incorporated into its state. Secondly, the base class provides a \texttt{register\_state} method for explicitly registering all other attribute types. In \ours, core components such as agents and memory inherit from \texttt{StateModule}. This design not only eliminates boilerplate code but also provides developers with \texttt{state\_dict} and \texttt{load\_state\_dict} methods for saving and restoring of the entire nested agent hierarchy.

\paragraph{Non-Invasive Customization.}
For high extensibility, we instrument the agent lifecycle with a comprehensive system of hooks, enabling developers to modify runtime behavior without altering the core codebase. Hooks are available as pre- and post-events at key operational points, including \texttt{reply}, \texttt{observe}, \texttt{reasoning}, \texttt{acting}, and the console output method, \texttt{print}.

Note that these hooks are not just passive listeners, they can actively modify the inputs and outputs of their respective functions. This capability supports a wide range of applications, from implementing detailed logging and validation rules to altering the agent’s reasoning path. For example, the \texttt{pre\_print} hook can intercept messages intended for the console and redirect them to a web-based user interface, effectively decoupling the agent’s core logic from its presentation layer.

\subsection{Built-in Agents}

\paragraph{Deep Research Agent.}
\begin{figure}[t]
    \centering
    \includegraphics[width=0.85\linewidth]{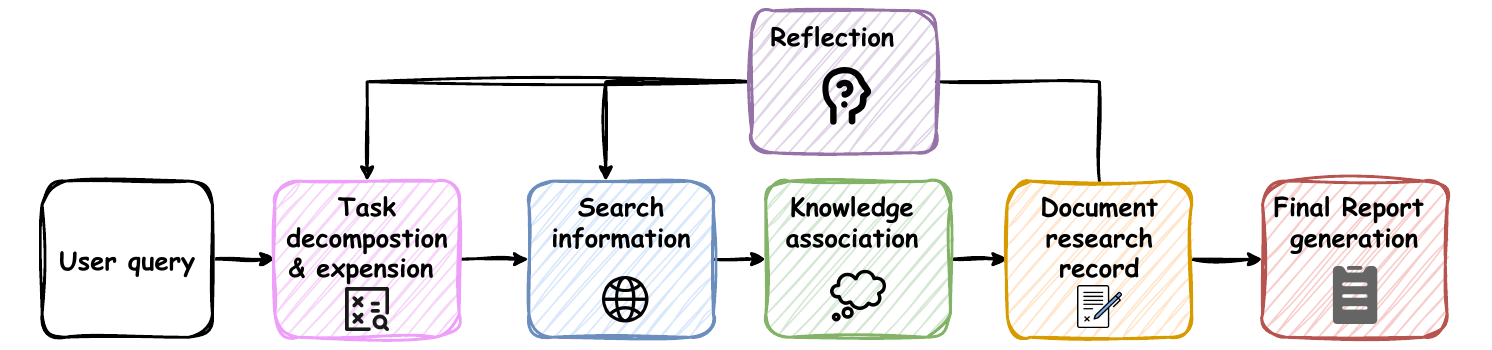}
    \caption{The workflow of Deep Research Agent.}
    \label{fig:deep_research_workflow}
\end{figure}
The Deep Research Agent is designed to search, gather, and combine information from multiple sources using search APIs, \eg Tavily MCP~\citep{tavily}, to provide report-formatted answers to users’ queries. It can generate detailed, well-organized reports that help users gain deeper insights towards the queried task. A workflow of Deep Research Agent is shown as \figref{fig:deep_research_workflow}.

The Deep Research Agent focuses on developing three core capabilities: query expansion, reflection, and summarization. These capabilities are abstracted into tools that the agent can invoke as needed. The process of query expansion involves continuously breaking down tasks into manageable sub-tasks, which transforms the linear workflow of the ReAct worker into a tree-based structure. During the search process, the agent conducts a broad reading by using multiple queries to explore a wide range of related knowledge, followed by a close reading where it extracts comprehensive content from select valuable web pages. If the information gathered is insufficient, the task is further decomposed into sub-tasks for deeper exploration.
The reflection capability of the Deep Research Agent is designed to address different types of failures with tailored strategies for trajectory optimization. Low-level reflection involves corrective measures for issues arising from tool errors, incorrect parameter usage, or ineffective sub-task completion. These are resolved by adjusting decision-making in subsequent steps of the ReAct process. On the other hand, high-level reflection addresses persistent failures that resist simple corrections, often indicating unanticipated practical challenges in the initial planning. In such cases, the agent may rephrase current steps if there is a misunderstanding of sub-task objectives or if they are unachievable in their current forms. For summarization, the agent mimics human research behavior by documenting useful results during the search process, forming a draft report without strict formatting requirements. This approach ensures that essential information is not overlooked, enabling the agent to proactively explore related topics from multiple perspectives and engage in in-depth reasoning, ultimately resulting in thorough analysis and comprehensive coverage of the subject matter. 

Another key strength of the Deep Research Agent is its integration with the Memory module in AgentScope. With this feature, the agent can store and revisit important information throughout its research process, further enhancing its ability to produce high-quality and comprehensive reports. 

\paragraph{Browser-use Agent.}

The Browser-use Agent is designed to autonomously navigate and interact with websites by integrating LLMs with browser automation tools such as Playwright MCP~\citep{playwright}. Typical applications encompass booking flights and hotels, querying stock prices and consolidating relevant news, web scraping and information summarization, submitting online forms, and monitoring real-time updates of specific web content, such as sports events or weather forecasts.

\begin{figure}[t]
    \centering
    \includegraphics[width=0.9\linewidth]{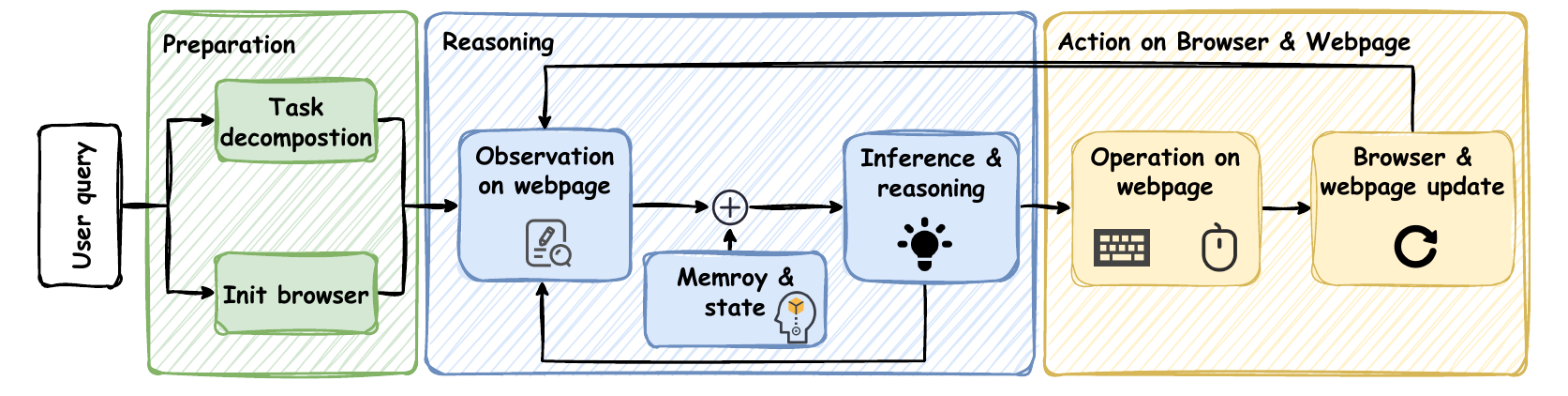}
    \caption{The workflow of Browser-user Agent.}
    \label{fig:browser_agent_workflow}
\end{figure}

An overview of the workflow of the Browser-use Agent is demonstrated in \figref{fig:browser_agent_workflow}. Key features and advantages of the Browser-use Agent include:
\begin{itemize}
    \item {\it Subtask Decomposition and Management}: The Browser-use Agent breaks down complex user queries into manageable subtasks, which it executes sequentially. This approach supports task updates and maintenance, enhancing the accomplish of tasks.
    \item {\it Integration of Visual and Web Textual Information}: By leveraging large models with visual capabilities, the Browser-use Agent is capable of reasoning over both webpage screenshots and HTML content, allowing for a deep understanding and more accurate interaction with various web pages.
    \item {\it Multi-Tab Browsing}: The Browser-use Agent supports concurrent management of multiple browser tabs, enabling parallel interactions with several web pages. This can be particularly helpful for workflows that require cross-referencing information and simultaneous monitoring.
    \item {\it Efficient Handling of Long Webpages}: To address the challenge of processing web pages that might exceed the context length limitation of LLMs, the Browser-use Agent segments long pages into smaller, manageable chunks. It performs webpage observation by chunk and manages cross-chunk contexts to ensure comprehensive information processing.    
\end{itemize}

With these abilities, the Browser-use Agent empowers users to efficiently gather information, perform complex interactions, and manage multiple subtasks, ultimately enabling them to solve complex problems through automatic navigation in web environments.

\paragraph{Meta Planner.}
Contemporary autonomous agent systems face significant challenges when tasked with complex, multi-step problems that require sophisticated planning, resource allocation, and coordination capabilities beyond the scope of traditional single-agent approaches.
While existing ReAct frameworks demonstrate proficiency in straightforward task execution through iterative reasoning-action cycles, they exhibit limitations when confronting intricate workflows that demand hierarchical task decomposition, specialized tool selection, and systematic progress tracking.
To address these constraints, we introduce the Meta Planner, a novel architectural agent that extends the ReAct paradigm through the integration of planning capabilities and dynamic worker orchestration. 
The system operates on a dual-mode architecture, automatically transitioning between lightweight ReAct processing for simple tasks and comprehensive planning-execution patterns for complex multi-stage problems, thereby optimizing computational resources while maintaining robust performance across diverse task complexities.

The Meta Planner implements a sophisticated planning-execution pipeline centered around three core functional modules: hierarchical task decomposition through structured roadmap generation, dynamic worker agent instantiation with specialized toolkit allocation, and persistent state management enabling long-term task continuity. 
The system employs a data structure for maintaining session context tracking.
Based on the session information data structure, the \texttt{RoadmapManager} module, as a set of tools, facilitates intelligent task breakdown into executable subtasks with defined dependencies and success criteria. 
Worker agents are dynamically created and managed through the tools provided in \texttt{WorkerManager} module, which allocates appropriate tool combinations—including MCP for external service integration-based on subtask requirements. 
This architecture enables the system to handle complex workflows such as multi-source data analysis, research synthesis, and iterative content generation, while maintaining transparency through comprehensive progress tracking and state persistence mechanisms that support task resumption and debugging capabilities.

\begin{figure}[h]
    \centering
    \includegraphics[width=0.85\linewidth]{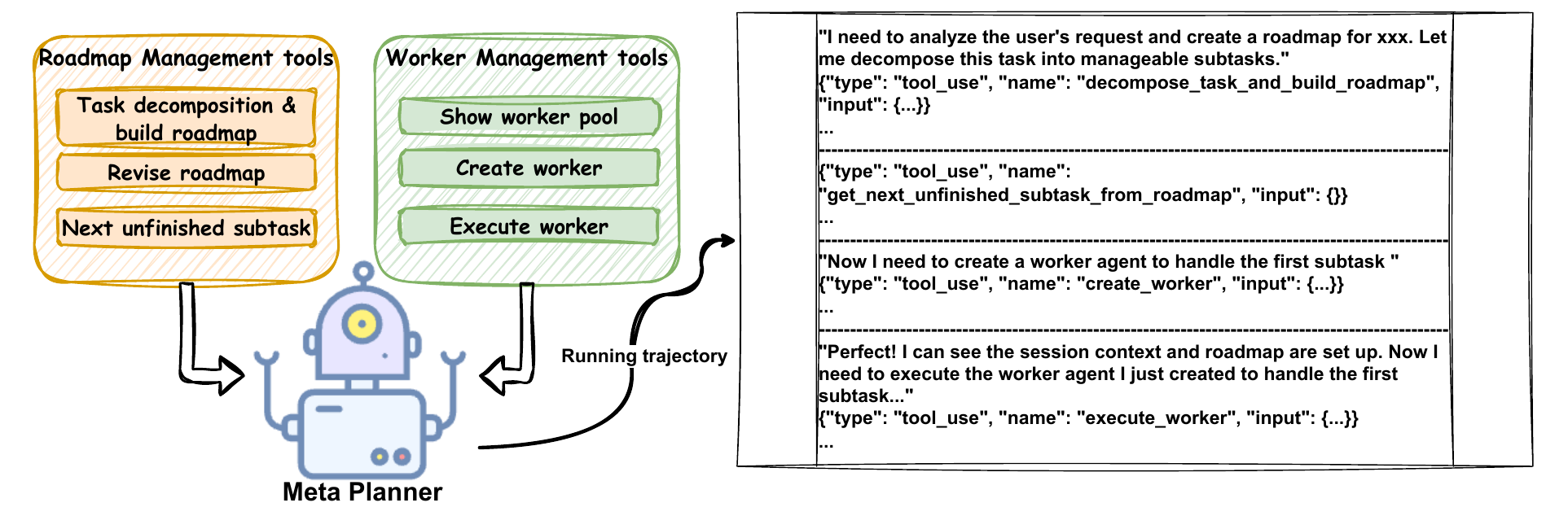}
    \caption{The key component of Meta Planner and an example of its trajectory.}
    \label{fig:meta_planner_agent}
\end{figure}

The agent features intelligent mode switching that automatically determines whether to use simple ReAct mode for straightforward tasks or advanced planning mode for complex multi-step operations.
An illustrative trajectory is provided in \figref{fig:meta_planner_agent}.

\subsection{Multi-Agent}
\label{subsec:multi-agent}

\subsubsection{Agent as a Tool}
\label{subsec:agent_as_tool}
In \ours, a widely used and recommended approach for building multi-agent applications is to treat agents as tools, \ie allowing agents to function as callable components within a large workflow. The intuition behind such an approach is that, while a primary agent still manages direct user interactions and conversations, it can autonomously select and invoke specialized agents as tools to handle particular subtasks or domains of expertise.

For example, as described in the recent study~\citep{li2025kimas}, a knowledge-integrated multi-agent system often requires different agents to manage distinct knowledge bases. When a user submits a query, the main agent routes the question to the appropriate agents (each instantiated as a tool and standing by to be called as needed). Upon receiving a request, these agents generate responses based on their knowledge bases. Finally, these outputs can be aggregated to deliver a comprehensive response to the user.

Such agent-as-a-tool architecture promotes scalability and flexibility of \ours. Agents can be independently developed, tested, and added to the system as new tools to rapidly adapt to evolving user requirements, enabling integration of novel capabilities or knowledge sources without disrupting existing workflows. 

\subsubsection{Agent Conversation}
\label{subsec:agent_conversation}

Agent conversation represents another standard paradigm for multi-agent applications. To streamline development and reduce complexity, \ours provides {\it pipelines} and {\it message hubs} for managing agent interactions efficiently and minimizing repetitive coding.

The pipeline abstraction encapsulates common patterns in agent conversation, including sequential, conditional, and iterative message exchanges, into simple and reusable components. Developers can construct agent conversations by assembling pipelines that handle the flow of messages between agents, enabling a clear separation between the interaction logic and the underlying message-passing mechanism. 
Pipelines can be employed in both functional and object-oriented styles, as shown in Example \ref{example:conversation_function}. Beyond basic sequential pipelines, AgentScope also offers constructs for conditional branching (\ie if-else and switch) and looped interactions (\ie while-loop and for-loop), making it easy to model complex and adaptive multi-agent behaviors.

\begin{lstlisting}[language={Python}, caption={Examples of chaining agents in a sequential manner.}, label={example:conversation_function}, float=t]
# 1: A functional implementation
from agentscope.pipeline import sequential_pipeline
msg = await sequential_pipeline(
    # List of agents to be executed in order
    agents=[alice, bob, charlie, david],
    # The first input message, can be None
    msg=None
)

# 2: A class-based implementation
from agentscope.pipeline import SequentialPipeline
# Create a pipeline object
pipeline = SequentialPipeline(agents=[alice, bob, charlie, david])
# Call the pipeline
msg = await pipeline(msg=None)
# Reuse the pipeline with different input
msg = await pipeline(msg=Msg("user", "Hello!", "user"))

\end{lstlisting}

The message hub abstraction acts as a centralized broadcast mechanism for simplifying group conversations among agents. By configuring a message hub with a set of participant agents and initial messages, developers can facilitate automatic message dissemination whenever any agent generates a new message, as illustrated in Example \ref{example:conversation_msghub}. The message hub ensures that all participating agents remain contextually synchronized and supports dynamic group dialogues~\citep{du2023improving}.

\begin{lstlisting}[language={Python}, caption={Broadcasting messagesa with message hub.}, label={example:conversation_msghub}]
async def example_broadcast_message():
    """Example of broadcasting messages with MsgHub."""

    # Create a message hub
    async with MsgHub(
        participants=[alice, bob, charlie],
        announcement=Msg(
            "user",
            "Now introduce yourself in one sentence, including your name, age and career.",
            "user",
        ),
    ) as hub:
        # Group chat without manual message passing
        await alice()
        await bob()
        await charlie()


asyncio.run(example_broadcast_message())
\end{lstlisting}

\section{Developer-Friendly Experience}
\label{sec:developer}
Towards developer-friendly experiences, we integrate comprehensive toolkits in \ours to further streamline the development, including {\it Evaluation}, {\it Studio}, and {\it Runtime}.

\subsection{Evaluation}
\label{sec:evaluation}

\subsubsection{From Tasks, Solutions and Metrics to Benchmark}
\begin{figure}[t]
    \centering
    \includegraphics[width=0.7\linewidth]{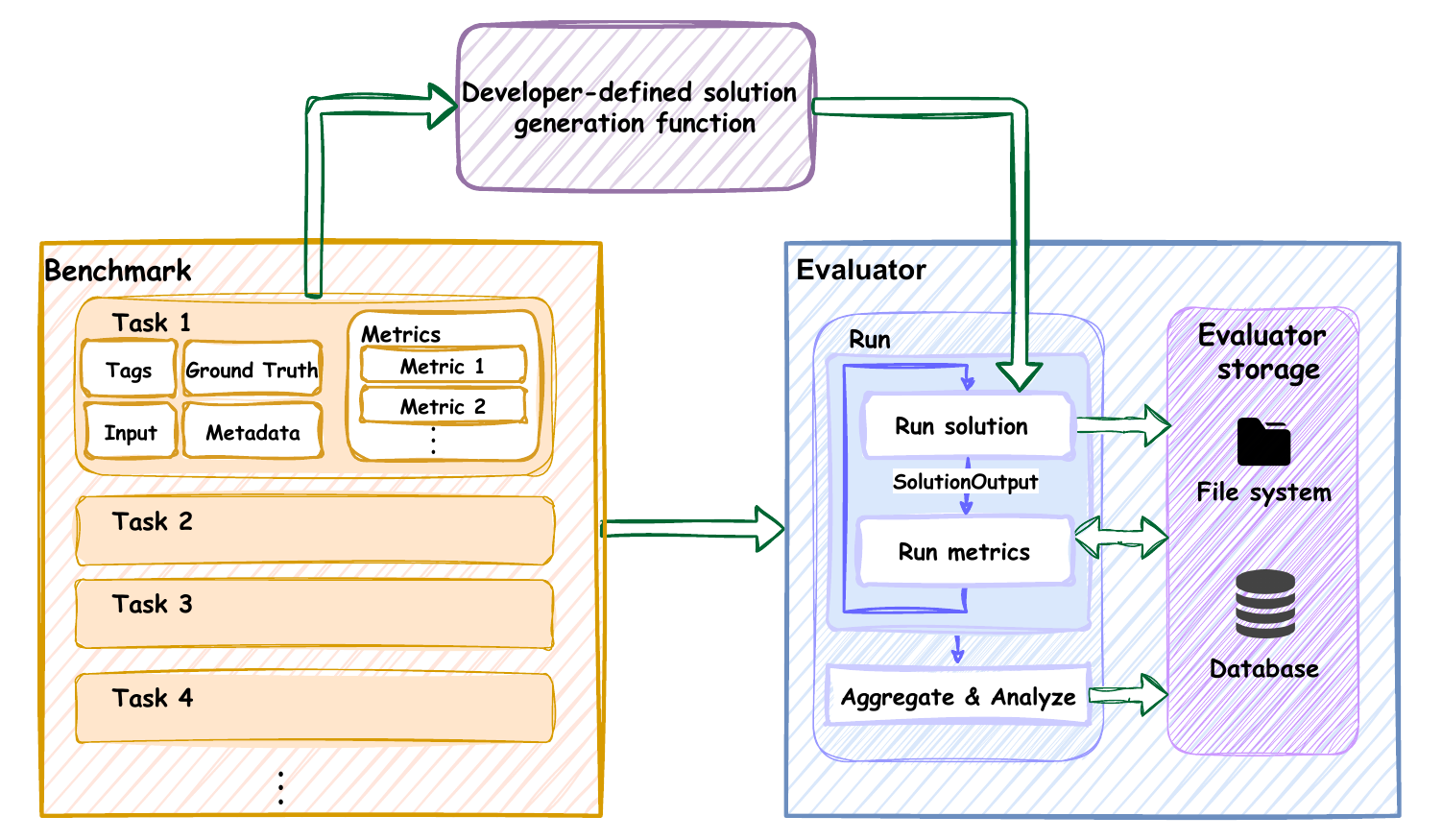}
    \caption{The evaluation module in \ours}
    \label{fig:evaluation}
\end{figure}

An overview of the evaluation module is illustrated in \figref{fig:evaluation}.
The evaluation module is designed with a hierarchical architecture that systematically organizes the several core components:

\begin{itemize}
    \item {\it Tasks}: A \texttt{Task} object represents an individual evaluation unit, encapsulating all the information required for agent execution and assessment. Each task is assigned a unique identifier and contains the task input, ground truth, evaluation metrics, and optional metadata such as category labels and additional context.
    
    \item {\it Solutions}: The evaluation framework defines a solution output class, \texttt{SolutionOutput}, to standardize the representation of agent-generated solutions. This structure captures three critical elements: (a) a success flag indicating whether the solution executed without exceptions, (b) the final output produced by the agent (\eg an answer or the terminal state of the environment), and (c) a complete trajectory documenting all tool callings and action results throughout execution. 
    This design enables both outcome-based and process-based evaluation approaches.
    
    \item {\it Metrics}: The abstract class \texttt{MetricBase} is implemented to support developer-defined metrics. The framework allows two primary metric types: categorical metrics, which yield discrete classifications (\eg \texttt{pass} or \texttt{fail}), and numerical metrics, which yield continuous scores. Each metric is a callable instance, and is expected to take a \texttt{SolutionOutput} object as input and generate a \texttt{MetricResult}.
    The generated \texttt{MetricResult} includes the metric name, computed score, timestamp, and optional messages for additional context. 
    This abstraction ensures the flexible integration of domain-specific evaluation criteria while maintaining consistency across various evaluation approaches.
    
    \item {\it Benchmarks}: A \textit{benchmark} aggregates multiple tasks into a cohesive evaluation suite by inheriting from \texttt{BenchmarkBase}. Benchmarks provide iterator functionality for systematic task traversal and implement indexing for random access patterns. This structure supports both sequential evaluation workflows and parallel processing strategies, allowing developers to construct domain-specific evaluation suites tailored to their experimental requirements.
\end{itemize}

\subsubsection{Evaluators}

To orchestrate the evaluation process, we integrate the \textit{Evaluator} module in \ours.
Developers can seamlessly transition between debugging-focused sequential evaluation and production-scale distributed evaluation without modifying their solution generation logic or benchmark definitions. 
The standardized interfaces are defined in \texttt{EvaluatorBase}, which manages evaluation across benchmark tasks and enables customization of application-specific evaluation pipelines. Two specific evaluators are provided, allowing users to prioritize either debugging capabilities or computational efficiency as needed.

On one hand, we implement a \texttt{GeneralEvaluator}, which executes tasks sequentially within a single process, making it particularly suited for development and debugging scenarios. 
This evaluator can be extended via user-defined solution generation functions that accept a task and a pre-hook callable as input, and return a coroutine producing a \texttt{SolutionOutput}. 
The sequential execution manner in \texttt{GeneralEvaluator} supports straightforward debugging, comprehensive logging, and step-by-step analysis of agent behavior.

On the other hand, we provide the \texttt{RayEvaluator} for high efficiency, leveraging the Ray~\citep{ray} distributed computing framework to enable parallel and distributed evaluation across multiple workers. 
This evaluator is designed for large-scale benchmark execution, where computational efficiency is essential. The Ray-based implementation automatically distributes tasks across available workers, manages resource allocation, and aggregates results, while maintaining the same interface as the sequential evaluator.

Both evaluators are integrated with the storage subsystem via \texttt{EvaluatorStorageBase}, which enables persistent storage of evaluation results, metadata, and experimental configurations. 
The framework supports evaluation continuation after interruptions by tracking completed tasks and resuming from appropriate checkpoints. Besides, we provide aggregation functionality in these evaluators for computing summary statistics across multiple task repetitions.

\subsection{Studio}
\label{sec:studio}

\begin{figure}[t]
    \centering
    \begin{subfigure}{0.85\linewidth}
        \centering
        \includegraphics[width=\linewidth]{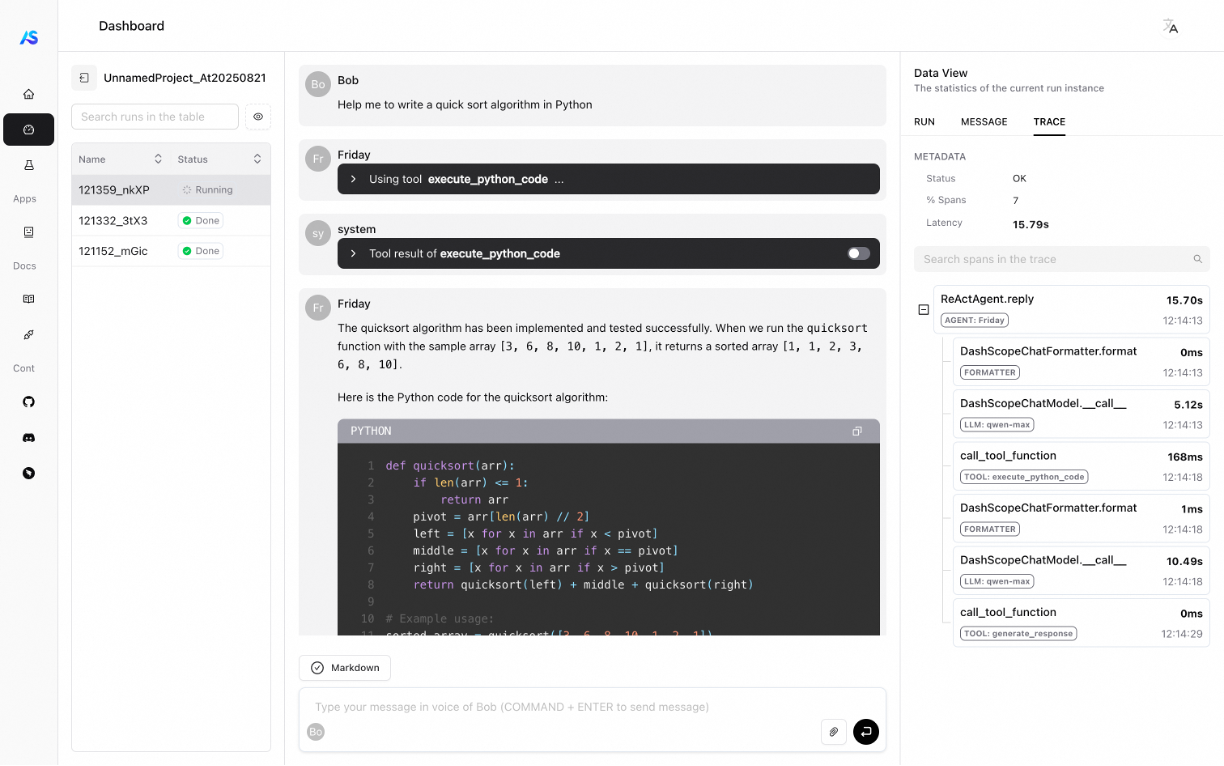}
        \caption{The chatbot-style dialogue and tracing visualization.}
        \label{fig:studio_tracing}
    \end{subfigure}
    \begin{subfigure}{0.85\linewidth}
        \centering
        \includegraphics[width=\linewidth]{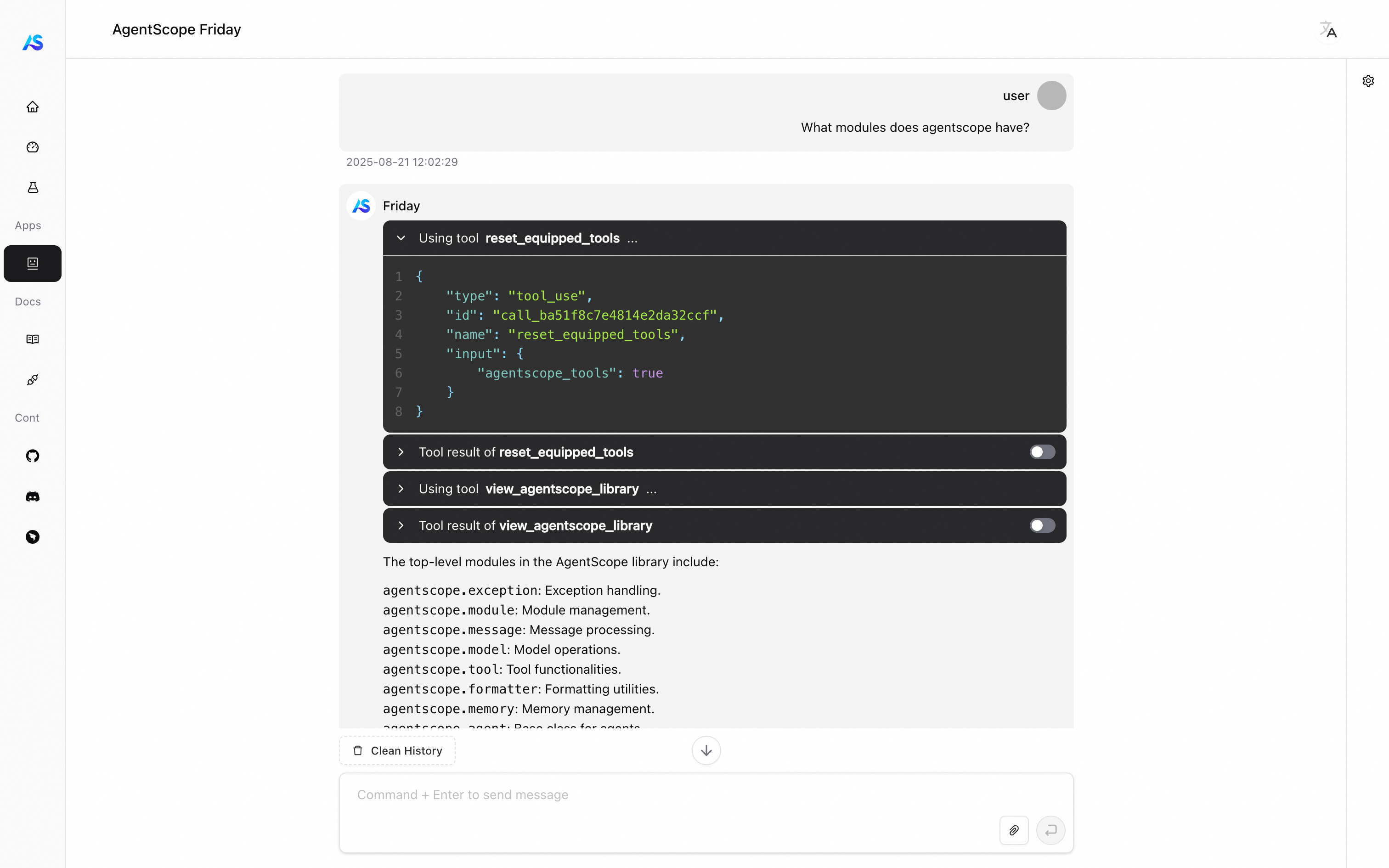}
        \caption{A built-in copilot Friday.}
        \label{fig:studio_tracing}
    \end{subfigure}
    \caption{Demonstrations of {\tt Studio} in \ours.}
    \label{fig:studio}
\end{figure}

The {\it Studio} in \ours serves as a visual platform designed to enhance transparency and control of the development of agentic applications. It is built upon a native integration with the OpenTelemetry standard~\citep{OpenTelemetry}, enabling the direct consumption and rendering of detailed telemetry data generated within applications.
Demonstrations of Studio are illustrated in~\figref{fig:studio}.

\subsubsection{Chatbot-style Dialogue and Tracing}

Developers can connect their applications to the Studio via a simple \texttt{init} function. 
Once connected, all messages, user inputs, and tracing data are streamed in real time to Studio’s web frontend, which visualizes agent interactions in an intuitive and chatbot-style dialogue interface. 
Such a dialogue view provides immediate clarity on the conversational flow by explicitly displaying structured message components, such as thinks, tool invocations, action results, and multimodal content.

Furthermore, the Studio also offers granular execution tracing for deep analysis. 
The execution trace depicts the entire operation as a hierarchical sequence of time-stamped spans, with each span representing a discrete computational step, \eg an LLM invocation, a tool execution, or the occurrence of an exception. 
Notably, each span in the trace view is directly linked to its associated message in the dialogue, allowing developers to seamlessly navigate from observed events in the conversation to underlying performance metrics in the trace. 
This tight integration enables rapid identification of latency sources, such as a slow LLM response or an inefficient tool function, accelerating the debugging and optimization process.

\subsubsection{Visualization of Evaluation Results}
\label{sec:eval_visual}

In collaboration with the evaluation module (refer to \secref{sec:evaluation}), Studio provides a dedicated visualization component that transforms raw evaluation results into interactive visualizations. 
By representing performance as a statistical distribution, developers are empowered to assess an agent’s stability and capabilities with greater statistical confidence, moving beyond simplistic and single-point metrics.

The visualization process begins with the ingestion of evaluation artifacts. When evaluation results are imported, Studio automatically parses and organizes them by their corresponding benchmarks, creating a structured foundation for further analysis. 
Rather than displaying a static value, agent performance is visualized as a comprehensive probability distribution. 
The visualization intelligently adapts to the type of metric, \eg discrete categories are shown differently from continuous metrics.
To ensure statistical validity, especially with a limited number of trials, Studio employs techniques such as bootstrapping~\cite{} to compute confidence intervals and render the full performance distribution. 
This approach offers a transparent and robust perspective on an agent’s stability and expected performance range, representing a significant improvement over potentially misleading averages.

Beyond high-level summaries, Studio diagnoses the sources of performance variation. 
It provides an aggregated statistical view that analyzes outcomes on a per-item basis across all trials, effectively grouping test items into cohorts such as "consistently correct", "consistently incorrect", or "unstable". 
This breakdown enables developers to quickly identify specific problem types where the agent excels or struggles, guiding optimization efforts toward the most impactful areas.

Studio supports trajectory comparisons for a fine-grained analysis. When the agent exhibits performance differences in the distribution tails, Studio allows side-by-side visual comparison of the corresponding execution trajectories. 
By juxtaposing both chains of tool calls, reasoning steps, and LLM responses, developers can conduct fine-grained root-cause analysis. This direct visual comparison makes it possible to pinpoint the exact divergence in the agent’s behavior that led to failure, effectively closing the loop from high-level statistical observations to actionable and low-level debugging insights.

\subsubsection{Built-in Copilot: Friday}
\label{sec:friday}

Studio includes a built-in copilot (\ie an assistant agent) named \textit{Friday}. This agent serves a dual purpose. On the one hand, it is designed to actively assist developers. On the other hand, it serves as a practical showcase of the advanced capabilities available in \ours, such as real-time steering (\secref{sec:realtime_steering}), dynamic tool provisioning (\secref{sec:agent}), and long-term memory management (\secref{sec:memory}).

Specifically, Friday is equipped with a specialized set of tools that grant it access to resources provided in \ours, \eg source code, tutorials, and documents, allowing it to search for technical information and generate helpful responses. 
In this way, Friday transforms the static documentation into a dynamic and conversational resource, providing immediate and context-aware assistance that accelerates both learning and development.
A developer can ask Friday to retrieve the exact signature of a function from the Python SDK or to find answers within the README and FAQs. 

Furthermore, as a showcase agent, Friday offers developers a concrete reference implementation. 
Instead of relying solely on abstract examples, users gain access to a live and feature-rich agent that demonstrates advanced usage patterns and facilitates a better understanding of the framework's capabilities.

\subsection{Runtime}
\label{subsec:runtime}

The deployment of agentic applications presents challenges in service orchestration, protocol compatibility, and secure tool execution. To tackle these challenges, we integrate \textit{\ASR}\footnote{\url{https://github.com/agentscope-ai/agentscope-runtime}} in \ours, a comprehensive agent runtime system designed for agent deployment and safe sandboxed tool execution.

Specifically, \ASR employs a dual-core architecture consisting of an \textit{Engine} and a \textit{Sandbox}. The Engine module provides the underlying infrastructure for deploying and managing agent applications, featuring built-in context management, session handling, and control over the tool sandbox. Meanwhile, the Sandbox module offers isolated environments to ensure secure tool execution. 

\paragraph{Engine.}
With the help of {\tt Engine} module, developers can create a {\tt Runner} object and pass an agent as one of its parameters. With applying the function {\it deploy}, the agent can be effortlessly deployed, automatically generating a production-ready FastAPI service with integrated health monitoring, graceful lifecycle management, and standardized API protocols. 
It is worth noting that \ours offers native support for multiple agent communication protocols, including Google’s Agent-to-Agent (A2A) protocol~\citep{a2a} and custom protocol adapters, ensuring seamless interoperability across heterogeneous agent ecosystems. Example \ref{example:a2a} provides an example of a deployment with A2A protocol support.

\begin{lstlisting}[language={python},caption={Examples of a deployment with A2A protocol.}, label={example:a2a}, float=t]
# Create and configure agent
agent = AgentScopeAgent(
    name="Friday",
    model=OpenAIChatModel("gpt-4"),
    agent_builder=ReActAgent,  # Or your agent class built with AgentScope
)

# Create executable runner
runner = Runner(
    agent=agent,
    context_manager=ContextManager(),
    environment_manager=EnvironmentManager(),
)

# Deploy as a production service with A2A protocol support
await runner.deploy(
    deploy_manager=LocalDeployManager(
        host="localhost",
        port=8090,
    ),
    endpoint_path="/process",
    protocol_adapters=A2AFastAPIDefaultAdapter(agent=agent),
)
\end{lstlisting}

\paragraph{Sandbox.}
The Sandbox provides a function-style interface that maintains consistent programming patterns while ensuring complete isolation. 
It supports various specialized environments (\eg \texttt{Filesystem} \texttt{Sandbox} for secure file operations, \texttt{BrowserSandbox} for web automation, and \texttt{TrainingSandbox} for benchmark evaluation) while maintaining consistent interfaces across different sandbox types. Developers can effortlessly extend their applications with additional MCP servers, without the overhead of preparing secure tool execution environments. 
An example of using the Sandbox module is shown in Example \ref{example:sandbox}.

\begin{lstlisting}[language={python}, caption={Examples of using the Sandbox module.}, label={example:sandbox}, float=t]
# Secure tool execution with automatic sandbox management
from agentscope_runtime.sandbox.tools.base import run_ipython_cell
result = run_ipython_cell(code="import os; print(os.listdir())")

# Persistent sandbox for stateful operations
with BaseSandbox() as sandbox:
    func = run_ipython_cell.bind(sandbox=sandbox)
    func(code="data = [1, 2,3]")
    # State preserved across calls
    func(code="print(sum(data))")
\end{lstlisting}

With \ASR, we deliver a developer-friendly experience that goes beyond deployment simplicity and backward compatibility with agentic applications. It also offers enhanced features such as structured communication protocols, multi-modal content support, and comprehensive lifecycle management. \ASR not only reduces deployment complexity, but also guarantees enterprise-grade reliability and security for agent applications, allowing developers to focus on agent logic instead of infrastructure concerns.

\section{Signature Applications}
\label{sec:application}
In this section, we introduce several signature applications of \ours, offering developers with hands-on tutorials from both implementation and execution.

\subsection{User-assistant Conversation}
In \exaref{example:react}, we demonstrate how to construct a user-assistant conversation by explicitly exchanging messages. 
The first step is to initialize both the ReAct agent and the user agent. For the ReAct agent, initialization involves specifying its name, system prompt, model, formatter, toolkit, and memory-related settings. The ReAct agent is built in an implementation-agnostic manner, with the main components exposed to the constructor, allowing developers to easily modify their agents without altering the core codebase. 
It is compatible with various model providers, including but not limited to OpenAI, DashScope, Gemini, Anthropic, and self-hosted open-source models.  

After the react agent and user agent are configured, the conversation can be built by having them exchange messages. In this setup, the ReAct agent and the user take turns speaking until the user decides to exit the interaction by typing the "exit" command.

\begin{lstlisting}[language={Python}, caption={An example of building a user-assistant conversation.}, label={example:react}, float=t]
import asyncio, os

from agentscope.agent import ReActAgent, UserAgent
from agentscope.formatter import DashScopeChatFormatter
from agentscope.memory import InMemoryMemory
from agentscope.model import DashScopeChatModel
from agentscope.tool import Toolkit, execute_shell_command, execute_python_code, view_text_file


async def main() -> None:
    """The main entry point for the ReAct agent example."""
    toolkit = Toolkit()
    toolkit.register_tool_function(execute_shell_command)
    toolkit.register_tool_function(execute_python_code)
    toolkit.register_tool_function(view_text_file)

    agent = ReActAgent(
        name="Friday",
        sys_prompt="You are a helpful assistant named Friday.",
        model=DashScopeChatModel(
            api_key=os.environ.get("DASHSCOPE_API_KEY"),
            model_name="qwen-max",
            enable_thinking=False,
            stream=True,
        ),
        formatter=DashScopeChatFormatter(),
        toolkit=toolkit,
        memory=InMemoryMemory(),
        # Additional arguments setting
        long_term_memory=Mem0LongTermMemory(),
        long_term_memory_mode="both",
        parallel_tool_call=True,
    )
    user = UserAgent("Bob")

    msg = None
    while True:
        msg = await user(msg)
        if msg.get_text_content() == "exit":
            break
        msg = await agent(msg)


asyncio.run(main())
\end{lstlisting}

\subsection{Multi-agent Conversation}
\label{sub:conversation}
\ours natively supports multi-agent conversations, primarily enabled by two key components: \texttt{MsgHub}, which manages message exchange among agents, and \texttt{Pipelines}, which orchestrate the interaction flow. 
These two components greatly simplify the development of complex conversational dynamics.

In \exaref{example:conversation}, we provide a practical demonstration of building a multi-agent conversation. 
The example begins by instantiating three agents, each with a distinct persona (\eg a teacher, a student, and a doctor). 
These agents are grouped within a \texttt{MsgHub}, which initiates the dialogue by broadcasting a system message that prompts each agent to introduce themselves. Then a \texttt{sequential\_pipeline} is used to ensure the agents speak in a predefined order.

To showcase dynamic group management, we remove the agent "Bob" from the \texttt{MsgHub}, announcing his departure to the remaining participants via a broadcast message. 
The example concludes by observing the reactions of the other agents, demonstrating the system’s ability to handle dynamic changes within a conversation.

\begin{lstlisting}[language={Python}, caption={An example of building a multi-agent conversation.}, label={example:conversation}, float=t]
import asyncio, os

from agentscope.agent import ReActAgent
from agentscope.formatter import DashScopeMultiAgentFormatter
from agentscope.message import Msg
from agentscope.model import DashScopeChatModel
from agentscope.pipeline import MsgHub, sequential_pipeline


def create_agent(name: str, age: int, career: str, character: str):
    """Create a participant agent with a specific name, age, and character."""
    return ReActAgent(
        name=name,
        sys_prompt=(
            f"You're a {age}-year-old {career} named {name} and you're "
            f"a {character} person."
        ),
        model=DashScopeChatModel(
            model_name="qwen-max",
            api_key=os.environ["DASHSCOPE_API_KEY"],
            stream=True,
        ),
        # Use multiagent formatter because multiple entities involves
        formatter=DashScopeMultiAgentFormatter(),
    )


async def main() -> None:
    """Run a multi-agent conversation workflow."""
    # Create multiple participant agents with different characteristics
    alice = create_agent("Alice", 30, "teacher", "friendly")
    bob = create_agent("Bob", 14, "student", "rebellious")
    charlie = create_agent("Charlie", 28, "doctor", "thoughtful")

    # Create a conversation where participants introduce themselves
    async with MsgHub(
        participants=[alice, bob, charlie],
        # The greeting message will be sent to all participants at the start
        announcement=Msg(
            "system",
            "Now you meet each other with a brief self-introduction.",
            "system",
        ),
    ) as hub:
        # Quick construct a pipeline to run the conversation
        await sequential_pipeline([alice, bob, charlie])

        # Delete a participant agent from the hub and fake a broadcast message
        hub.delete(bob)
        await hub.broadcast(
            Msg(
                "bob",
                "I have to start my homework now, see you later!",
                "assistant",
            ),
        )
        await alice()
        await charlie()

asyncio.run(main())
\end{lstlisting}

\subsection{Deep Research Agent}

The Deep Research Agent\footnote{The implementation details of the Deep Research Agent can be found at \url{https://github.com/agentscope-ai/agentscope/tree/main/examples/agent\_deep\_research}.} extends the ReAct agent with specialized research methodologies designed to handle complex queries, excelling at data collection, comprehensive investigation, and synthesis.
The agent initialization establishes a connection to a Tavily search service through MCP integration~\citep{tavily}, providing powerful web search and content extraction capabilities.

During execution, one can observe that the agent automatically breaks down research questions into manageable subtasks, conducts targeted searches for each component, identifies knowledge gaps that require further investigation, and synthesizes findings into coherent reports.
The agent maintains intermediate memory for tracking research progress and can generate structured outputs including detailed analysis reports, making it particularly suitable for academic research, market analysis, technical investigations, and comprehensive fact-finding missions that require multi-source verification and deep analytical reasoning.

\subsection{Browser-use Agent}

The Browser-use Agent\footnote{The implementation details of the Browser-use Agent can be found at \url{https://github.com/agentscope-ai/agentscope/tree/main/examples/agent_browser}.} extends the ReAct agent with specialized browser capabilities via the PlayWright MCP~\citep{playwright}, which provides essential browser operation tools.

The initialization begins by establishing a stateful connection through the \texttt{StdIOStatefulClient}, which communicates with the MCP server using standard input/output protocols. These tools are then registered to a toolkit by integrating the stateful client. The Browser Agent is configured with some specific components, including the model, formatter, memory, and the browser-enabled toolkit, while other parameters are inherited from the ReAct Agent.

The agent automatically manages browser states using specialized functions that support task decomposition, subtask manager, screenshot taking, chunk-wise webpage observation, memory summarization, and tool execution result filtering. 
During each interaction cycle, it captures webpage snapshots (and screenshots if the LLM has vision ability), reasons about the current browser state, and executes appropriate actions such as navigation, clicking, and typing.

\begin{figure}[t]
    \centering
    \includegraphics[width=0.9\linewidth]{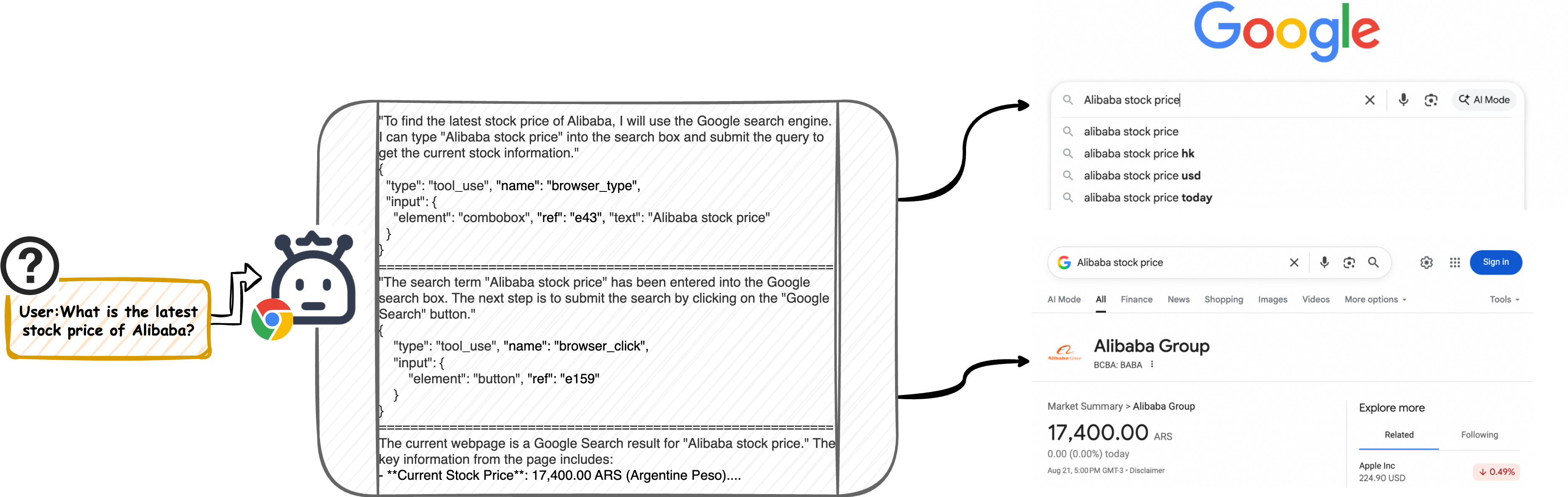}
    \caption{A running demonstration of the built-in browser-use agent.}
    \label{fig:browser_agent}
\end{figure}

Through its conversational loop, users can naturally issue web automation commands, such as "search for Python tutorials" or "navigate to GitHub and find trending repositories", while the agent handles the complex sequences of browser interactions required to fulfill these requests. \figref{fig:browser_agent} shows a screenshot of the agent responding to the query "What is the latest stock price of Alibaba?", where it successfully finds the relevant information via Google search in the browser.

\subsection{Meta Planner}

The Meta Planner\footnote{The implementation details of the Meta Planner can be found at \url{https://github.com/agentscope-ai/agentscope/tree/main/examples/meta_planner_agent}.} extends the ReAct agent with advanced planning mechanisms that decompose complex tasks into manageable subtasks and orchestrate specialized worker agents for efficient completion.

During initialization, the agent establishes two distinct toolkits: a \textit{planner} toolkit for high-level planning operations, and a \textit{worker} toolkit equipped with comprehensive tools, including shell command execution, file operations, web search, and filesystem access through MCP clients. Multiple MCP clients are configured to provide external tool integration, allowing the agent to access search functionality and manage filesystem operations within a designated working directory.

The Meta Planner operates on a planning-execution pattern with three core components:  (a) A dataset structure containing roadmap information for managing session context and user inputs (refer to \exaref{example:metaplanner_roadmap});  (b) A set of \textit{roadmap management} tools for task decomposition and progress tracking;  (c) \textit{Worker management} tools for creating and supervising specialized worker agents.

State persistence is built into the agent’s workflow, automatically saving progress at key stages, including post-reasoning and post-action states, which supports recovery from interruptions and the resumption of long-running tasks, thereby simplifying debugging during extended sessions.
These capabilities make the meta planner especially well-suited for complex workflows such as comprehensive data analysis, research projects, content creation, and sophisticated problem-solving tasks that require coordinated execution across multiple domains.

\begin{lstlisting}[caption={An example of a roadmap generated by the meta planner.}, label={example:metaplanner_roadmap}, float=t]
{
    ...,
    "roadmap": {
        "original_task": "Create a comprehensive analysis report of Meta (META) stock that includes a company overview and key financial metrics from Q1 2025, with particular focus on profit margins.",
        "decomposed_tasks": [
            {
                "subtask_specification": {
                    "subtask_description": "Research and gather comprehensive company overview information about Meta Platforms Inc.",
                    "input_intro": "Need to collect current information about Meta's business operations, market position, and recent developments",
                    "exact_input": "Research Meta Platforms Inc. (META) - gather information about: business model and main revenue streams, recent major developments and strategic initiatives, market position in social media/metaverse space, current leadership and corporate structure, main products and services (Facebook, Instagram, WhatsApp, Reality Labs, etc.)",
                    "expected_output": "A comprehensive company overview document containing Meta's business model, recent developments, market position, leadership, and main products/services",
                    "desired_auxiliary_tools": "tavily-search for current company information and recent news"
                },
                "status": "Planned",
                "updates": [],
                "attempt": 0,
                "workers": []
            },
            ...
        ]
    },
}
\end{lstlisting}

\section{Conclusions}
We introduce \ours 1.0, a flexible and extensible framework that leverages the ReAct paradigm to integrate reasoning and action for LLM-based agents. This integration facilitates seamless interaction between agents and their environments through dynamic tool use. By incorporating modular foundational agent components, efficient agent-level infrastructure, and customizable interfaces, \ours provides a robust solution that bridges the gap between prototype agents and real-world applications. The framework also features a suite of developer-friendly toolkits, which simplify the development process and enhance the usability and flexibility of agentic applications.
Looking ahead, we envision \ours as a practical foundation for building scalable, adaptive, and trustworthy agentic systems. By supporting tool-based perception and interaction, \ours effectively addresses the evolving demands of LLM-based applications, equipping agents to tackle increasingly complex real-world tasks with autonomy and precision.

\bibliography{citation}
\bibliographystyle{colm2024_conference}

\end{document}